\documentclass[11pt]{article}

\usepackage{acl}

\usepackage{times}
\usepackage{latexsym}

\usepackage[T1]{fontenc}

\usepackage[utf8]{inputenc}

\usepackage{microtype}

\usepackage{inconsolata}

\usepackage{graphicx}
\usepackage{stfloats}
\usepackage{enumitem}
\usepackage{booktabs}
\usepackage{multirow}
\usepackage{graphicx}

\usepackage{xcolor}
\usepackage{hyperref}
\usepackage[most]{tcolorbox}
\usepackage{enumitem}
\usepackage[ruled,vlined,linesnumbered]{algorithm2e}
\usepackage{float}
\usepackage{caption}
\usepackage{placeins}

\newcounter{prompt}[section]
\renewcommand{\theprompt}{\thesection.\arabic{prompt}}

\newtcolorbox{promptbox}[2][]{
    enhanced,
    colback=gray!10,
    colframe=gray!50,
    coltitle=white,
    colbacktitle=gray!60,
    title={Prompt~\theprompt: #2},
    fonttitle=\bfseries,
    boxrule=0.4pt,
    arc=2pt,
    width=\columnwidth,
    left=4pt,
    right=4pt,
    top=4pt,
    bottom=4pt,
    breakable,
    #1
}

\title{DKCD: Domain Knowledge-Enhanced Causal Discovery from Unstructured Data}

\author{
Xin Li \and Jin Li \and Shoujin Wang \and Kun Yu \and Fang Chen \\
University of Technology Sydney \\
Sydney, NSW, Australia \\
\texttt{xin.li-19@student.uts.edu.au} \\
\texttt{jin.li-4@student.uts.edu.au} \\
\texttt{shoujin.wang@uts.edu.au} \\
\texttt{Kun.Yu@uts.edu.au} \\
\texttt{Fang.Chen@uts.edu.au}
}

\begin{document}
\maketitle
\begin{abstract}
Causal discovery from unstructured data is a challenging yet underexplored task in high-expertise domains such as healthcare, finance, and education. Existing methods typically leverage the general knowledge of large language models (LLMs) to identify causal factors from unstructured data and annotate them into structured data for causal graph construction. However, they remain limited by two key challenges (CHs): \textbf{(CH1) insufficient identification of latent factors}, which are implicit in the data yet essential for causal discovery, due to the lack of domain-specific knowledge; and \textbf{(CH2) unreliable factor annotation}, caused by the lack of domain-grounded reasoning, which propagates errors to the resulting causal graphs. To address these challenges, we introduce a novel \textbf{D}omain \textbf{K}nowledge-enhanced \textbf{C}ausal \textbf{D}iscovery framework (DKCD) for causal discovery from unstructured data in high-expertise domains with three interconnected components: \textbf{(1) Knowledge Mining:} It retrieves relevant domain knowledge based on observable factors to support subsequent causal reasoning. \textbf{(2) Knowledge-guided Causal Reasoning:} Reasoning with relevant knowledge, it discovers latent causal factors to address CH1 and generates key causal clues for more accurate data annotation to address CH2. and \textbf{(3) Causal Structure Discovery:} It constructs the final causal graphs based on a more complete factor set and accurate annotations. Experiments on two domain-specific datasets show that DKCD significantly improves both causal factor identification and causal graph construction. 
\end{abstract}

\section{Introduction}

\begin{figure*}[t]
    \centering
    \includegraphics[width=\linewidth]{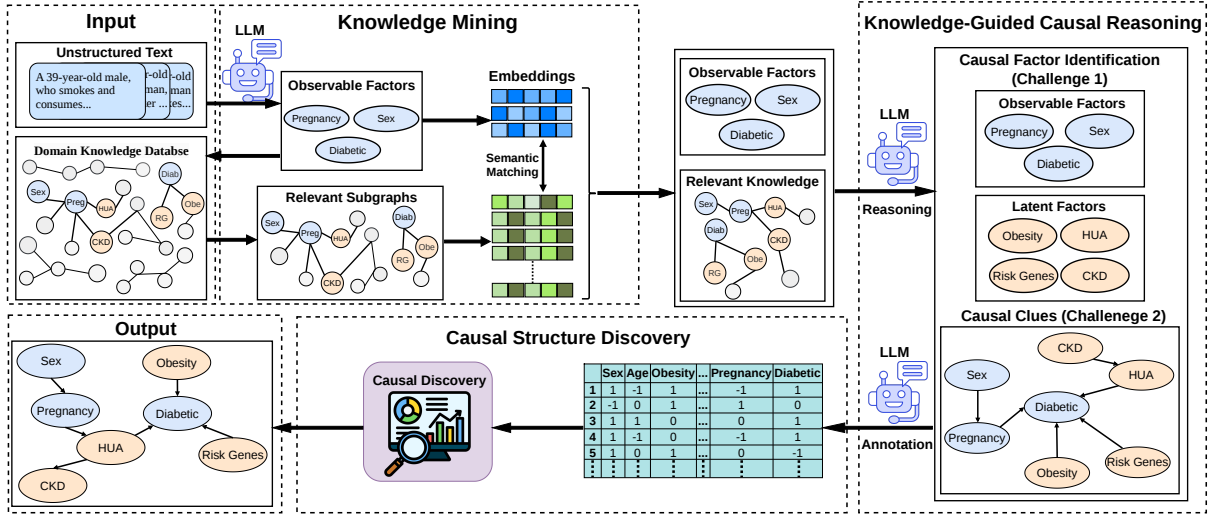}
    \caption{Overview of the DKCD framework. DKCD leverages domain knowledge graphs to guide latent factor discovery and enhances LLM-based causal reasoning by injecting external domain knowledge. The resulting knowledge-grounded factor scoring improves the robustness and accuracy of downstream causal graph construction.}
    \label{fig:DKCD}
    \vspace{-16pt}
\end{figure*}

Exploring and uncovering causality from real-world data is fundamental across a wide range of scientific domains \cite{bunge2017causality, illari2011causality}, including healthcare \cite{shi2022learning, yang2013causal}, finance \cite{papana2017financial, sokolov2025toward}, and education \cite{morrison2016searching, fancsali2014causal}. In data-driven causal discovery, causal relationships are typically represented as causal graphs.  Existing causal discovery algorithms, such as the Peter–Clark (PC) algorithm \cite{spirtes2000causation, ramsey2016improving} and the Fast Causal Inference (FCI) algorithm \cite{spirtes2013causal}, are primarily designed for structured data with predefined schemes and fixed formats, commonly organized in tabular form (e.g., relational databases and spreadsheets). Thus, they face substantial limitations when applied to unstructured data \cite{malinsky2018causal, yu2016review}. Meanwhile, the growing dominance of unstructured data, particularly natural-language text \cite{siddiqa2017big, eberendu2016unstructured, azad2020role}, calls for methods that bridge unstructured data and statistical causal discovery.

With the emergence of large language models (LLMs), new opportunities have arisen for analyzing and discovering causal relationships from unstructured data \cite{wang2023causal, wang2024causalbench}. With strong capabilities for understanding, reasoning, and extracting knowledge from natural language, LLMs can support automatic identification of causal factors, reducing reliance on manual domain expertise for data labeling~\cite{ashwani2024cause, dubey2024llama}. Specifically, the COAT framework \cite{liu2024discovery} is a representative approach that leverages LLMs to extract causal factors from unstructured data, annotate factor values, and then apply statistical causal discovery algorithms to construct causal graphs. Although COAT takes an important step forward, it suffers from critical limitations when applied to high-expertise domains where causal factor identification requires substantial domain knowledge, such as clinical text analysis, where many causal factors are implicit rather than explicitly stated in unstructured data.

\begin{figure}[t]
    \centering
    \includegraphics[width=0.5\textwidth]{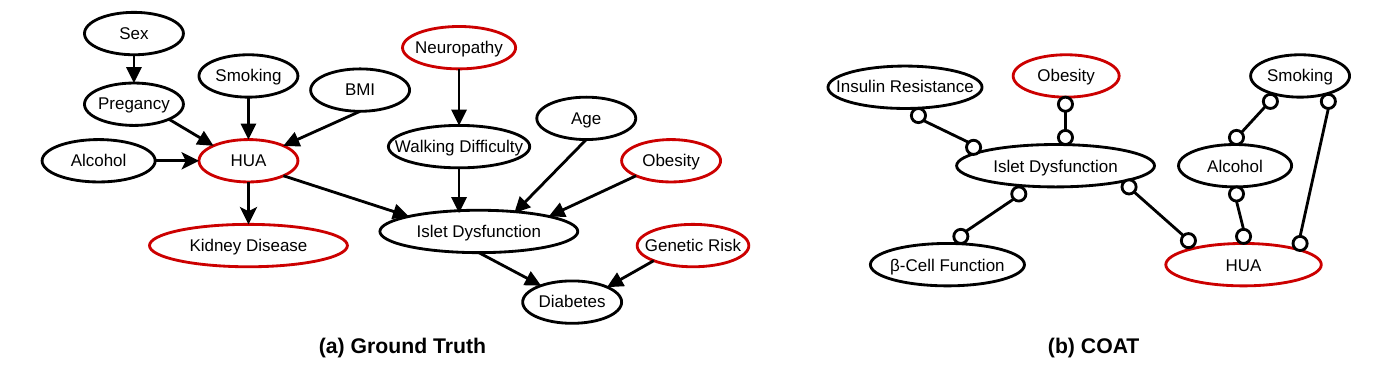}
    \vspace{-10pt}
    \caption{The COAT result on the diabetes dataset.}
    \label{fig:IntroFigExper}
    \vspace{-16pt}
\end{figure}

As shown in Figure~\ref{fig:IntroFigExper}, applying COAT to a diabetes-domain dataset reveals limitations from two key challenges (CHs). \textbf{Challenge 1 (CH1):} COAT relies solely on LLMs' general knowledge to propose causal factors. Although such knowledge is broad, it often lacks domain knowledge \cite{yu2024onsep, shou2023pairwise}, making it difficult to identify latent factors implied by domain-specific observations. For instance, COAT misses kidney disease, highlighted as a red node, which requires knowledge linking reduced kidney function, albuminuria, and diabetes-related complications to renal conditions. Such omissions yield an incomplete factor set and limit the final causal graph. \textbf{Challenge 2 (CH2):} Causal graph construction also requires accurate factor-value annotations as structured input to downstream statistical causal discovery algorithms. However, without external domain knowledge, LLMs' limited causal reasoning capabilities \cite{mirzadeh2024gsm} often produce inaccurate or inconsistent annotations \cite{burford2024generative, zevcevic2023causal, ban2025integrating}. For example, COAT may annotate HUA mainly based on explicit factors such as alcohol or smoking while missing the role of kidney disease, causing errors that propagate into the inferred the final causal graph as missing causal relationships and uncertain or incorrect edge directions. Although Li et al. \cite{li2025revealing} extended COAT \cite{liu2024discovery} to multimodal datasets through MLLM-CD, the same challenges remain.

To address these challenges, we propose the DKCD (\textbf{D}omain \textbf{K}nowledge-enhanced \textbf{C}ausal \textbf{D}iscovery) framework, which effectively leverages domain knowledge graphs (KGs) to guide both causal factor discovery and annotation, thereby improving the quality of statistically inferred causal graphs from unstructured data. As illustrated in Figure~\ref{fig:DKCD}, DKCD consists of three components. (1) Knowledge Mining: It extracts observable factors from unstructured data using LLMs and retrieves relevant subgraphs from domain KGs. Semantic matching filters the retrieved knowledge to preserve the most relevant context, supporting subsequent causal reasoning. (2) Knowledge-Guided Causal Reasoning: Building on the observable factors and retrieved knowledge context, this module infers latent factors that are not explicitly stated in the unstructured data but are causally relevant in the domain, thereby directly addressing the incomplete factor identification challenge (\textbf{CH1}). In addition, the module generates \emph{causal clues}, i.e., domain-grounded descriptions of plausible dependencies among factors. These clues provide structured guidance for scoring and annotation, helping reduce ambiguity and improve annotation reliability, thereby addressing the inaccurate factor annotation challenge (\textbf{CH2}). (3) Causal Structure Discovery: With a more complete factor set and more accurate annotations, this module applies a statistical causal discovery algorithm to infer the final causal graph. In addition, Figure~\ref{fig:ExampleDKCD} in Appendix illustrates DKCD's causal discovery process using a Diabetes dataset example. The main contributions of our work are briefly summarized as follows:
\begin{itemize}[leftmargin=*, itemsep=2pt, topsep=2pt, parsep=0pt, partopsep=2pt]
    \item We propose DKCD, a novel domain knowledge-enhanced causal discovery framework that integrates domain KGs with LLM-based reasoning to enhance causal discovery from unstructured data in high-expertise domains.
    \item We design a knowledge-guided causal reasoning module that leverages retrieved domain knowledge to discover latent causal factors and generate knowledge-grounded causal clues, thereby improving annotation accuracy.
    \item We construct the first domain-specific benchmarks with real-world knowledge graphs for evaluation. Extensive experiments show that DKCD significantly outperforms existing methods regarding both causal factor identification and causal graph construction.
\end{itemize}

\section{Related Work}

\textbf{Causal Discovery} aims to identify causal relationships from data and typically represents them as causal graphs. Existing methods are commonly divided into three categories: constraint-based \cite{niu2024comprehensive, vowels2022d}, score-based \cite{vowels2022d}, and continuous optimization methods \cite{gong2024causal, hyttinen2016causal}. Constraint-based methods infer structure through conditional independence tests, with representative algorithms including FCI \cite{spirtes2013causal}, PC \cite{spirtes2000causation, ramsey2016improving}, and PCMCI \cite{runge2019detecting}. Score-based methods evaluate candidate graphs using criteria such as the Bayesian Information Criterion (BIC) and the Bayesian Dirichlet equivalent uniform (BDeu) \cite{huang2018generalized, nogueira2022methods}, while continuous optimization methods reformulate structure learning as differentiable optimization, such as NOTEARS \cite{zhou2022survey, lopez2022large}. Although existing causal discovery algorithms perform well on structured data, they are difficult to apply directly to unstructured text. DKCD bridges this gap by enabling effective causal discovery on domain-specific unstructured data.

\noindent
\textbf{LLM for Causal Discovery} has recently emerged as a promising direction \cite{gopalakrishnan2024causality, wu2023bloomberggpt, tong2024automating}. Prior studies use LLMs to extract causal factors \cite{liu2024discovery}, estimate causal edges \cite{ban2023causal, darvariu2024large}, and support counterfactual reasoning from natural language data \cite{gendron2024counterfactual, liu2026crest}. However, LLMs alone do not reliably perform causal reasoning, as their predictions are largely driven by pretraining data, which limits robustness in domain-specific settings.

\noindent
\textbf{KG-Guided LLMs for Causal Discovery} introduce structured domain knowledge to enhance LLM-based causal reasoning and causal analysis \cite{kim2024causal, pan2024unifying}. Representative studies include RC2R \cite{yu2024fusing}, which focuses on causal reasoning over existing knowledge graphs, and RealTCD \cite{li2024realtcd}, which performs causal discovery on structured time-series data, using textual information only as auxiliary knowledge. Other studies incorporate biomedical knowledge graphs into prompting or retrieval frameworks to support LLM reasoning \cite{susanti2024knowledge, lin2026knowledge}. Despite these advances, LLMs still struggle to reliably distinguish correlation from causation \cite{wu2024causality}, particularly in specialized domains where spurious correlations are common. Moreover, these are not directly applicable to our setting, which requires domain knowledge to extract potential causal factors and reliable annotations from unstructured data.

\section{Domain Knowledge-Enhanced Causal Discovery}

\subsection{Problem Definition}

We consider a dataset $\mathcal{D} = \{\mathbf{x}_1, \mathbf{x}_2, \dots, \mathbf{x}_n\}$, where each sample $\mathbf{x}_k \in \mathcal{X}$ represents a domain-specific textual description. Examples of the generated patient condition descriptions used in our experiments are provided in Appendix~\ref{appendix:dataset_examples}. In domain-specific settings such as healthcare, textual descriptions often contain rich unstructured observations that can support the identification of potential causal factors and relationships \cite{li2025can, ban2025integrating}. Let $\mathcal{V}^* = \{V_1, V_2, \dots, V_d\}$ denote the complete set of ground-truth causal factors underlying the domain process. The ground-truth causal relationships among these variables are represented by a directed acyclic graph (DAG) $G^* = (\mathcal{V}^*, E^*)$, where $E^*$ denotes the set of directed causal edges. Therefore, causal discovery from unstructured data aims to infer a causal graph from $\mathcal{D}$ that approximates the ground-truth structure $G^*$.


\noindent
\textbf{Identifying Causal Factors in Specialized Domains.} This task aims to identify a relevant set of causal factors $\mathcal{V}$ from an unstructured dataset $\mathcal{D}$. However, unstructured data often provides only partial observations of the underlying causal factors and omits several important causal factors, either because they require domain expertise to infer or because they are implicitly reflected through related symptoms, conditions, or measurements. To better formulate this problem, we divide the causal factors to be identified as $\mathcal{V}=\mathcal{V}_o\cup\mathcal{V}_l$. Here, $\mathcal{V}_o$ denotes \textit{observable causal factors}, which are directly mentioned in the observational data and can therefore be easily extracted by existing methods like COAT. $\mathcal{V}_l$ denotes \textit{latent causal factors}, which are not explicitly stated but essential to obtain the correct and comprehensive causal graph. For example, given diabetic patient descriptions, factors such as neuropathy, hyperuricemia (HUA), kidney disease, obesity, and genetic predisposition may not be explicitly mentioned, but they are clinically important for diabetes diagnosis. They can be inferred with domain-knowledge-grounded reasoning from indirect clinical indicators, such as kidney disease, which may be reflected by abnormal eGFR or albuminuria. However, existing methods commonly overlook them (e.g., Figure~\ref{fig:IntroFigExper}) due to the lack of capabilities to obtain and leverage domain-specific knowledge. Therefore, this motivates the knowledge-guided design of DKCD.

\noindent
\textbf{Annotating Factor Values.} Given the identified factors $\mathcal{V}$, each text $\mathbf{x}_k$ must be annotated with a value $v_{kj}$ for each variable $V_j \in \mathcal{V}$, converting $\mathcal{D}$ into a structured scoring table
{
\setlength{\abovedisplayskip}{4pt}
\setlength{\abovedisplayshortskip}{4pt}
\setlength{\belowdisplayskip}{4pt}
\setlength{\belowdisplayshortskip}{4pt}
\begin{equation}
\mathbf{S} = \{(v_{k1}, v_{k2}, \cdots, v_{k|\mathcal{V}|})\}_{k=1}^{n},
\label{eq:scoring_table}
\end{equation}
}

\noindent
where $|\mathcal{V}|$ denotes the cardinality of $\mathcal{V}$. Existing methods \cite{li2025revealing, liu2024discovery} typically obtain $v_{kj}$ by directly prompting LLMs to annotate the value of each factor for each sample. This annotation step is critical because the quality of $\mathbf{S}$ directly determines the accuracy of downstream causal discovery. However, accurately scoring factors, especially latent ones, in unstructured data requires domain-grounded reasoning, which LLMs lack when operating without external knowledge.

\noindent
\textbf{Discovering the Causal Structure.} Given the structured data $\mathbf{S}$, a statistical causal discovery algorithm is applied to infer the causal graph $G$. The accuracy of the inferred graph depends on both the completeness of the factor set and the reliability of the annotations. Errors in either distort conditional independence tests and propagate to the graph.

\noindent
\textbf{The proposed DKCD framework} aims to address these challenging tasks by integrating LLMs with domain knowledge graphs and statistical causal discovery algorithms for reliable causal structure discovery. The overall framework is illustrated in Figure~\ref{fig:DKCD} and consists of three key components. First, the knowledge mining module extracts observable causal factors from unstructured text using LLMs and retrieves relevant subgraphs from the domain knowledge graph to support the subsequent causal reasoning. Second, the knowledge-guided causal reasoning leverages knowledge to discover latent causal factors and derive causal clues, thereby supporting more reliable annotation. Third, the causal structure discovery module applies statistical causal discovery algorithms to infer causal relationships among variables and construct the final causal graph. In addition, Figure~\ref{fig:ExampleDKCD} provides an illustrative example from the Diabetes dataset to demonstrate the overall process of the DKCD, which is further summarized in Algorithm~\ref{section:algorithm}.

\subsection{Knowledge Mining}
As illustrated in Figure~\ref{fig:DKCD}, this module takes a domain-specific textual description and the domain KG, denoted by $\mathcal{G}_{KG}$, as input. The objective of this module is not only to extract directly observable causal variables, but also to retrieve the most relevant contextual knowledge for subsequent knowledge-guided reasoning.

\noindent
\textbf{Observable Factor Proposal:} Given a small set of textual samples $\mathcal{X}_s = \{\mathbf{x}_1, \mathbf{x}_2, \dots, \mathbf{x}_m\}$, where $\mathcal{X}_s \subset \mathcal{D}$ and each sample $\mathbf{x}_k \in \mathcal{X}_s$ denotes a textual description, the LLM is prompted (see Prompt~\ref{prompt:observable_mining} in Appendix~\ref{appendix:prompts}) to identify a set of observable causal factors $\mathcal{V}_o$ explicitly mentioned in the unstructured data.
This step is formulated as the following factor extraction mapping:
{
\setlength{\abovedisplayskip}{1pt}
\setlength{\abovedisplayshortskip}{1pt}
\setlength{\belowdisplayskip}{1pt}
\setlength{\belowdisplayshortskip}{1pt}
\begin{equation}
f_{\text{obs}}:\mathcal{X}_s \rightarrow \mathcal{V}_o.
\end{equation}
}

\noindent
These factors provide the initial semantic anchors for subsequent retrieval of relevant knowledge and latent factor mining. However, these identifying factors alone may still miss important latent causal factors, motivating the use of domain KGs for latent causal factor identification. 

\noindent
\textbf{Domain KGs Retrieval:}
Based on the identified observable factors $\mathcal{V}_o$, we retrieve relevant subgraphs from the domain KG, denoted as $\mathcal{G}_{KG}$. Specifically, the observable factors are first normalized without altering their semantic meaning, then matched to relevant entity nodes in $\mathcal{G}_{KG}$. Their neighboring entities and relations are subsequently collected to construct contextual subgraphs
{
\setlength{\abovedisplayskip}{1pt}
\setlength{\abovedisplayshortskip}{1pt}
\setlength{\belowdisplayskip}{1pt}
\setlength{\belowdisplayshortskip}{1pt}
\begin{equation}
\mathcal{G}_{rel}^{(n)} \subseteq \mathcal{G}_{KG},
\end{equation}
}

\noindent
where $\mathcal{G}_{rel}^{(n)}$ denotes the retrieved subgraph with $n$ triples, where $n$ is the number of retrieved triples. In the semantic matching stage, these triples are further transformed into natural language statements for relevance ranking. The retrieved subgraph contains structured domain concepts relevant to the observable factors. The domain KG provides external knowledge to enrich textual causal semantics and support latent factor identification.

\noindent
\textbf{Semantic Matching:}
To perform finer-grained retrieval aligned with semantic information, we further conduct semantic matching between observable factors and retrieved knowledge entities to preserve the most relevant knowledge context. Since pretrained sentence encoders are trained on natural language rather than symbolic triples, the retrieved triples are verbalized into natural language statements for semantic similarity matching \cite{galal2024rethinking, yin2024study}. Specifically, let $\mathcal{S}^{(k)} = \{s_1, s_2, \dots, s_{n_k}\}$ denote the set of natural language statements verbalized from the triples in the retrieved subgraph. This transformation aligns the structured KG information with the textual embedding space used by pretrained sentence encoders, enabling more effective semantic matching with patient condition descriptions. For each observable factor $v_i \in \mathcal{V}_o$, we encode $v_i$ and each statement $s_j \in \mathcal{S}^{(k)}$ into dense embeddings using the pretrained sentence-transformer model all-MiniLM-L6-v2 \cite{yin2024study, vergou2023readability}, denoted as $E(\cdot)$, and compute their semantic similarity using cosine similarity:
{
\setlength{\abovedisplayskip}{3pt}
\setlength{\abovedisplayshortskip}{3pt}
\setlength{\belowdisplayskip}{3pt}
\setlength{\belowdisplayshortskip}{3pt}
\begin{equation}
\mathrm{sim}(v_i, s_j) = \frac{E(v_i)^\top E(s_j)}{\|E(v_i)\|\|E(s_j)\|}.
\end{equation}
}

\noindent
We then select the top-$r$ most relevant statements for each observable factor:
{
\setlength{\abovedisplayskip}{2pt}
\setlength{\abovedisplayshortskip}{2pt}
\setlength{\belowdisplayskip}{2pt}
\setlength{\belowdisplayshortskip}{2pt}
\begin{equation}
\mathcal{S}_{v_i}^{(k)} = \operatorname{Top}\text{-}r \{ s_j \in \mathcal{S}^{(k)} \mid \mathrm{sim}(v_i, s_j) \},
\end{equation}
}

\noindent
where $\mathcal{S}_{v_i}^{(k)}$ denotes the set of matched knowledge statements associated with factor $v_i$. Further analysis of the parameter $r$ is provided in Appendix~\ref{appendix:parameter}. The selected knowledge items are then used to construct the relevant knowledge context for each sample, providing more reliable support for latent factor discovery and causal clue inference.

\subsection{Knowledge-Guided Causal Reasoning}
This module leverages the retrieved domain knowledge to perform two key reasoning tasks that directly address the two identified challenges. First, it infers latent causal factors absent from the surface text, thereby addressing the problem of incomplete factor identification. Second, it generates knowledge-grounded causal clues that guide more reliable and accurate factor annotation.

\noindent
\textbf{Latent Factor Discovery:}
Given the observable factor set $\mathcal{V}_o$ and the matched knowledge context derived from $\mathcal{G}_{rel}^{(n)}$, the LLM performs knowledge-guided reasoning to infer a set of latent causal factors $\mathcal{V}_l$. These latent factors are not directly mentioned in the unstructured data, but are supported by the semantic relations and domain concepts provided by the retrieved knowledge. The complete causal factor set is therefore defined as
{
\setlength{\abovedisplayskip}{1pt}
\setlength{\abovedisplayshortskip}{1pt}
\setlength{\belowdisplayskip}{1pt}
\setlength{\belowdisplayshortskip}{1pt}
\begin{equation}
\mathcal{V} = \mathcal{V}_o \cup \mathcal{V}_l.
\end{equation}
}

\noindent
By incorporating domain knowledge to discover latent factors, this step extends causal factor discovery beyond surface-level textual mentions, leading to a more comprehensive factor set that better approximates the ground-truth factor set $\mathcal{V}^*$.

\noindent
\textbf{Causal Clue Generation:} Beyond factor discovery, the LLM derives causal clues from the interaction between the identified factors and the retrieved domain knowledge through internal reasoning ability. These causal clues describe plausible causal dependencies and domain-informed influence patterns among factors. Formally, for each sample $\mathbf{x}_k$, the generated causal clues are denoted as
{
\setlength{\abovedisplayskip}{2pt}
\setlength{\abovedisplayshortskip}{2pt}
\setlength{\belowdisplayskip}{2pt}
\setlength{\belowdisplayshortskip}{2pt}
\begin{equation}
\mathcal{H}^{(k)} = L(\mathcal{V}, \mathcal{G}_{rel}^{(k)}),
\end{equation}
}

\noindent
where $L(\cdot)$ represents the knowledge-guided reasoning process implemented by the LLM following the prompt \ref{prompt:causal_clue} shown in Appendix~\ref{appendix:prompts}. An example of the derived causal clues is shown in Appendix~\ref{appendix:causal_clue_example}. Note that the generated clues do not directly determine the final causal graph. Instead, they serve as domain-grounded guidance during the annotation step: by providing the LLM with explicit reasoning about how factors relate causally, the clues help produce more accurate and consistent factor values in the scoring table $\mathbf{S}$, thereby improving the reliability of the structured data fed into the downstream causal structure discovery module.

\subsection{Causal Structure Discovery}

With a more complete causal factor set $\mathcal{V}$ and more reliable annotations in the structured scoring table $\mathbf{S}$, the final step applies a statistical causal discovery algorithm $\mathcal{C}$ to infer the causal graph. Formally, the causal discovery process can be expressed as
{
\setlength{\abovedisplayskip}{2pt}
\setlength{\abovedisplayshortskip}{2pt}
\setlength{\belowdisplayskip}{2pt}
\setlength{\belowdisplayshortskip}{2pt}
\begin{equation}
G = \mathcal{C}\left(\mathbf{S}, \mathcal{V} \right),
\end{equation}
}

\noindent
where $\mathcal{C}$ denotes the causal discovery algorithm, and $G$ denotes the inferred causal graph whose nodes correspond to the variables in $\mathcal{V}$ and whose edges represent the discovered causal relationships. Moreover, we instantiate $\mathcal{C}$ using the FCI algorithm \cite{spirtes2013causal} for final causal structure discovery. FCI is particularly suitable because it can infer causal structures in the presence of latent confounders without requiring all causal factors to be observed~\cite{ramsey2025efficient, varghese2024causal}, which is important in specialized domains where some causal factors may remain unidentified. Since traditional causal discovery algorithms, including FCI, require structured tabular data and are sensitive to noisy or incomplete inputs \cite{malinsky2018causal}, the upstream modules of DKCD are designed to produce a more complete factor set and more accurate annotations, helping recover a causal graph closer to the ground truth $G^*$.

\section{Experiments}
We evaluate the performance of the DKCD framework in both causal factor discovery and causal structure discovery on synthetic datasets from the diabetes and respiratory domains, using multiple LLMs, including GPT-4o \cite{hurst2024gpt}, Gemini 2.5 Flash \cite{comanici2025gemini}, Grok-3 \cite{xai2025grok3}, and LLaMA 3-70B \cite{meta2024llama3}. Due to space constraints, more detailed descriptions of the dataset construction process, prompt design, implementation details, evaluation metrics, and complete experimental results, as well as further analyses of the ablation studies and parameter analysis, are provided in Appendix~\ref{sec:more_experiment_details}.

\begin{table*}[t]
\centering
\small
\setlength{\tabcolsep}{4pt}
\caption{Causal factor identification and structure discovery performance on the diabetes dataset.}
\label{tab:DiaResults}

\resizebox{\textwidth}{!}{
\begin{tabular}{llccccccc}
\toprule
LLM & Method & NP$\uparrow$ & NR$\uparrow$ & NF$\uparrow$ & AP$\uparrow$ & AR$\uparrow$ & AF$\uparrow$ & ESHD$\downarrow$ \\
\midrule

\multirow{5}{*}{GPT-4o}
& Zero-shot LLM & 0.26 $\pm$ 0.12 & 0.14 $\pm$ 0.07 & 0.15 $\pm$ 0.03 & 0.17 $\pm$ 0.29 & 0.03 $\pm$ 0.05 & 0.05 $\pm$ 0.08 & 36.67 $\pm$ 2.31 \\
& META & 0.57 $\pm$ 0.07 & 0.26 $\pm$ 0.11 & 0.35 $\pm$ 0.12 & \textbf{1.00 $\pm$ 0.00} & 0.13 $\pm$ 0.09 & 0.22 $\pm$ 0.13 & 32.00 $\pm$ 1.00 \\
& COAT & 0.61 $\pm$ 0.11 & 0.31 $\pm$ 0.09 & 0.41 $\pm$ 0.10 & 0.65 $\pm$ 0.04 & 0.18 $\pm$ 0.04 & 0.28 $\pm$ 0.05 & 30.67 $\pm$ 2.08 \\
& DKCD & \textbf{0.84 $\pm$ 0.16} & \textbf{0.71 $\pm$ 0.08} &\textbf{0.77 $\pm$ 0.04} & 0.50 $\pm$ 0.17 & \textbf{0.33 $\pm$ 0.18} & \textbf{0.40 $\pm$ 0.18} & \textbf{25.33 $\pm$ 3.21} \\

\midrule

\multirow{5}{*}{Gemini 2.5 Flash}
& Zero-shot LLM & 0.35 $\pm$ 0.07 & 0.17 $\pm$ 0.04 & 0.23 $\pm$ 0.06 & 0.50 $\pm$ 0.00 & 0.08 $\pm$ 0.00 & 0.14 $\pm$ 0.00 & 36.00 $\pm$ 3.61 \\
& META & 0.59 $\pm$ 0.08 & 0.31 $\pm$ 0.09 & 0.34 $\pm$ 0.09 & 0.81 $\pm$ 0.17 & 0.23 $\pm$ 0.08 & 0.36 $\pm$ 0.11 & \textbf{29.67 $\pm$ 1.53} \\
& COAT & 0.52 $\pm$ 0.25 & 0.24 $\pm$ 0.05 & 0.32 $\pm$ 0.09 & 0.37 $\pm$ 0.32 & 0.10 $\pm$ 0.12 & 0.16 $\pm$ 0.17 & 35.00 $\pm$ 7.00 \\
& DKCD & \textbf{0.67 $\pm$ 0.03} & \textbf{0.62 $\pm$ 0.25} & \textbf{0.59 $\pm$ 0.21} & \textbf{0.86 $\pm$ 0.15} & \textbf{0.33 $\pm$ 0.16} & \textbf{0.45 $\pm$ 0.17} & \textbf{29.67 $\pm$ 1.53} \\

\midrule

\multirow{5}{*}{Grok-3}
& Zero-shot LLM & 0.54 $\pm$ 0.27 & 0.16 $\pm$ 0.04 & 0.23 $\pm$ 0.06 & 0.17 $\pm$ 0.29 & 0.03 $\pm$ 0.05 & 0.05 $\pm$ 0.08 & 37.33 $\pm$ 2.08 \\
& META & \textbf{0.78 $\pm$ 0.13} & 0.41 $\pm$ 0.08 & 0.54 $\pm$ 0.09 & 0.81 $\pm$ 0.17 & 0.28 $\pm$ 0.05 & 0.41 $\pm$ 0.06 & \textbf{27.00 $\pm$ 4.58} \\
& COAT & 0.55 $\pm$ 0.04 & 0.26 $\pm$ 0.05 & 0.35 $\pm$ 0.05 & \textbf{0.83 $\pm$ 0.29} & 0.10 $\pm$ 0.04 & 0.18 $\pm$ 0.07 & 33.00 $\pm$ 0.00 \\
& DKCD & 0.71 $\pm$ 0.04 & \textbf{0.64 $\pm$ 0.07} & \textbf{0.67 $\pm$ 0.05} & 0.74 $\pm$ 0.14 & \textbf{0.41 $\pm$ 0.16} & \textbf{0.44 $\pm$ 0.12} & 30.33 $\pm$ 3.21 \\

\midrule

\multirow{5}{*}{LLaMA 3-70B}
& Zero-shot LLM & 0.56 $\pm$ 0.22 & 0.38 $\pm$ 0.18 & 0.45 $\pm$ 0.20 & 0.20 $\pm$ 0.09 & 0.29 $\pm$ 0.09 & 0.29 $\pm$ 0.09 & 29.67 $\pm$ 8.14 \\
& META & \textbf{0.79 $\pm$ 0.10} & 0.41 $\pm$ 0.04 & 0.53 $\pm$ 0.04 & 0.48 $\pm$ 0.14 & 0.20 $\pm$ 0.05 & 0.29 $\pm$ 0.07 & 39.00 $\pm$ 3.46 \\
& COAT & 0.74 $\pm$ 0.07 & 0.38 $\pm$ 0.08 & 0.50 $\pm$ 0.07 & 0.60 $\pm$ 0.35 & 0.18 $\pm$ 0.05 & 0.27 $\pm$ 0.09 & 29.00 $\pm$ 3.61 \\
& DKCD & 0.71 $\pm$ 0.06 & \textbf{0.57 $\pm$ 0.00} & \textbf{0.63 $\pm$ 0.07} & \textbf{0.65 $\pm$ 0.19} & \textbf{0.33 $\pm$ 0.12} & \textbf{0.44 $\pm$ 0.14} & \textbf{27.67 $\pm$ 3.51} \\

\bottomrule
\end{tabular}
}
\end{table*}

\begin{table*}[t]
\centering
\small
\setlength{\tabcolsep}{4pt}
\caption{Causal factor identification and structure discovery performance on the respiratory dataset.}
\label{tab:ResResults}

\resizebox{\textwidth}{!}{
\begin{tabular}{llccccccc}
\toprule
LLM & Method & NP$\uparrow$ & NR$\uparrow$ & NF$\uparrow$ & AP$\uparrow$ & AR$\uparrow$ & AF$\uparrow$ & ESHD$\downarrow$ \\
\midrule

\multirow{5}{*}{GPT-4o}
& Zero-shot LLM & 0.42 $\pm$ 0.07 & 0.38 $\pm$ 0.00 & 0.40 $\pm$ 0.03 & 0.44 $\pm$ 0.10 & 0.13 $\pm$ 0.00 & 0.20 $\pm$ 0.01 & 20.00 $\pm$ 5.20 \\
& META & 0.33 $\pm$ 0.06 & 0.25 $\pm$ 0.00 & 0.28 $\pm$ 0.02 & \textbf{1.00 $\pm$ 0.00} & 0.13 $\pm$ 0.00 & 0.23 $\pm$ 0.00 & 25.67 $\pm$ 1.53 \\
& COAT & 0.34 $\pm$ 0.06 & 0.25 $\pm$ 0.00 & 0.29 $\pm$ 0.02 & 0.67 $\pm$ 0.58 & 0.09 $\pm$ 0.08 & 0.15 $\pm$ 0.13 & 27.00 $\pm$ 4.00 \\
& DKCD & \textbf{0.74 $\pm$ 0.13} & \textbf{0.75 $\pm$ 0.00} & \textbf{0.74 $\pm$ 0.07} & 0.87 $\pm$ 0.23 & \textbf{0.56 $\pm$ 0.22} & \textbf{0.68 $\pm$ 0.21} & \textbf{16.67 $\pm$ 2.31} \\

\midrule

\multirow{5}{*}{Gemini 2.5 Flash}
& Zero-shot LLM & 0.39 $\pm$ 0.05 & 0.29 $\pm$ 0.08 & 0.33 $\pm$ 0.06 & \textbf{0.44 $\pm$ 0.51} &0.09 $\pm$ 0.08 & 0.14 $\pm$ 0.12 & 23.00 $\pm$ 6.24 \\
& META & 0.21 $\pm$ 0.06 & 0.13 $\pm$ 0.00 & 0.16 $\pm$ 0.02 & 0.33 $\pm$ 0.58 & 0.04 $\pm$ 0.08 & 0.08 $\pm$ 0.13 & 26.00 $\pm$ 4.36 \\
& COAT & 0.23 $\pm$ 0.06 & 0.21 $\pm$ 0.07 & 0.22 $\pm$ 0.07 & 0.33 $\pm$ 0.58 & 0.04 $\pm$ 0.08 & 0.08 $\pm$ 0.13 & 30.00 $\pm$ 1.73 \\
& DKCD & \textbf{0.84 $\pm$ 0.08} & \textbf{0.88 $\pm$ 0.13} & \textbf{0.86 $\pm$ 0.10} & 0.41 $\pm$ 0.16 & \textbf{0.42 $\pm$ 0.19} & \textbf{0.41 $\pm$ 0.17} & \textbf{17.33 $\pm$ 3.21} \\

\midrule

\multirow{5}{*}{Grok-3}
& Zero-shot LLM & 0.51 $\pm$ 0.09 & 0.38 $\pm$ 0.00 & 0.41 $\pm$ 0.02 & 0.44 $\pm$ 0.10 & 0.13 $\pm$ 0.00 & 0.20 $\pm$ 0.01 & 22.33 $\pm$ 3.21 \\
& META & 0.37 $\pm$ 0.04 & 0.29 $\pm$ 0.08 & 0.33 $\pm$ 0.05 & \textbf{1.00 $\pm$ 0.00} & 0.13 $\pm$ 0.00 & 0.23 $\pm$ 0.00 & 25.67 $\pm$ 2.08 \\
& COAT & 0.32 $\pm$ 0.18 & 0.21 $\pm$ 0.07 & 0.25 $\pm$ 0.11 & 0.33 $\pm$ 0.58 & 0.04 $\pm$ 0.08 & 0.08 $\pm$ 0.13 & 26.33 $\pm$ 6.03 \\
& DKCD & \textbf{0.67 $\pm$ 0.14} & \textbf{0.63 $\pm$ 0.21} & \textbf{0.64 $\pm$ 0.18} & 0.62 $\pm$ 0.13 & \textbf{0.34 $\pm$ 0.08} & \textbf{0.43 $\pm$ 0.09} & \textbf{18.67 $\pm$ 1.15} \\

\midrule

\multirow{5}{*}{LLaMA 3-70B}
& Zero-shot LLM & 0.32 $\pm$ 0.02 & 0.38 $\pm$ 0.00 & 0.34 $\pm$ 0.01 & 0.50 $\pm$ 0.00 & 0.13 $\pm$ 0.00 & 0.21 $\pm$ 0.00 & 27.00 $\pm$ 2.00 \\
& META & 0.35 $\pm$ 0.07 & 0.29 $\pm$ 0.08 & 0.32 $\pm$ 0.07 & \textbf{1.00 $\pm$ 0.00} & 0.13 $\pm$ 0.00 & 0.21 $\pm$ 0.00 & 25.67 $\pm$ 0.58 \\
& COAT & 0.53 $\pm$ 0.12 & 0.42 $\pm$ 0.07 & 0.47 $\pm$ 0.09 & 0.83 $\pm$ 0.29 & 0.25 $\pm$ 0.00 & 0.38 $\pm$ 0.04 & 23.33 $\pm$ 3.06 \\
& DKCD & \textbf{0.73 $\pm$ 0.09} & \textbf{0.80 $\pm$ 0.14} & \textbf{0.69 $\pm$ 0.11} & 0.81 $\pm$ 0.17 & \textbf{0.50 $\pm$ 0.25} & \textbf{0.57 $\pm$ 0.18} & \textbf{17.33 $\pm$ 5.03} \\

\bottomrule
\end{tabular}
}
\vspace{-5pt}
\end{table*}

\subsection{Experimental Setup}
Currently, there are no publicly available datasets that simultaneously provide unstructured data and aligned ground-truth causal graphs in domain-specific settings. Since causal discovery requires ground-truth causal graphs for quantitative evaluation, synthetic datasets remain necessary in existing research practice \cite{liu2024discovery, li2025revealing}. Following prior work, we construct two medical-domain synthetic datasets grounded in real-world medical KGs for evaluation: (1) the \textbf{Diabetic Patient Condition Descriptions Dataset (Diabetes Dataset)}, which contains 400 patient condition descriptions with 14 high-level factors and a diabetes-domain KG; and (2) the \textbf{Respiratory Patient Condition Descriptions Dataset (Respiratory Dataset)}, which contains 400 patient condition descriptions with 8 high-level factors and a respiratory-domain KG. Additional details are provided in Appendix~\ref{appendix:Implementation}.

\noindent
\textbf{Diabetes Dataset:} Following prior work \cite{liu2024discovery, li2025revealing}, we generate a medical dataset containing clinical descriptions of diabetic patients. Each synthetic sample contains a textual description of patient conditions and clinical information. This includes 9 observable factors (i.e., sex, pregnancy, alcohol, smoking, BMI, walking difficulty, age, diabetes, and islet dysfunction), and 5 latent factors (HUA, kidney disease, neuropathy, obesity, and genetic risk), as shown in Figure~\ref{fig:DiabetesCG}(a). Additionally, we construct a diabetes-domain KG using the DiaKG dataset \cite{chang2021diakg}. Additional construction details and the example of the generated data are provided in Appendix~\ref{appendix:diabetesConstruct}.

\noindent
\textbf{Respiratory Dataset:} We also generate another medical dataset containing clinical descriptions of patients with respiratory conditions, following prior work \cite{liu2024discovery, li2025revealing}. Each sample includes 5 observable factors (i.e., tuberculosis, smoke, lung cancer, lung disease, and chest X-ray) and 3 latent factors (i.e., Asian travel, bronchitis, and dyspnea), as illustrated in Figure~\ref{fig:RespiratoryCG_GPT}(a). Additionally, we construct a respiratory-domain KG based on prior studies. Additional construction details and the example of the generated data are provided in Appendix~\ref{appendix:RespiratoryConstruct}.

\noindent
\textbf{Baselines:} Since causal discovery directly from unstructured data remains underexplored, we include all relevant and comparable baselines in the literature. Zero-shot LLM \cite{antonucci2023zero, du2025causal} generates causal graphs directly using LLMs without domain knowledge. COAT \cite{liu2024discovery} is a state-of-the-art LLM-driven framework that combines LLM with traditional causal discovery algorithms. META \cite{liu2024discovery} is a variant derived from the COAT ablation study. Other related methods are not included because they focus on different settings, such as causal reasoning over existing KGs \cite{yu2024fusing} or causal discovery from structured data \cite{li2024realtcd}, rather than directly discovering causal graphs from unstructured data. Additional details are provided in Appendix~\ref{appendix:baselines}.

\noindent
\textbf{Metrics:} Following common practice in the causal discovery literature, we evaluate the performance of factor identification and graph structure using widely adopted metrics \cite{zanga2022survey, li2025revealing, liu2024discovery}. For causal factor identification, we report: (1) Node Precision (NP), (2) Node Recall (NR), and (3) Node F1-score (NF). For causal structure discovery, we adopt: (1) Adjacency Precision (AP), (2) Adjacency Recall (AR) and (3) Adjacency F1-score (AF). To jointly assess both aspects, we also employ an extended Structural Hamming Distance (ESHD) \cite{peters2015structural, li2025revealing}. All results are reported as the mean and standard deviation over three runs. Additional details are provided in Appendix~\ref{appendix:metrics}.

\subsection{Analysis on the Datasets}
\begin{figure*}[t]
    \centering
    \vspace{-10pt}
    \includegraphics[width=\linewidth]{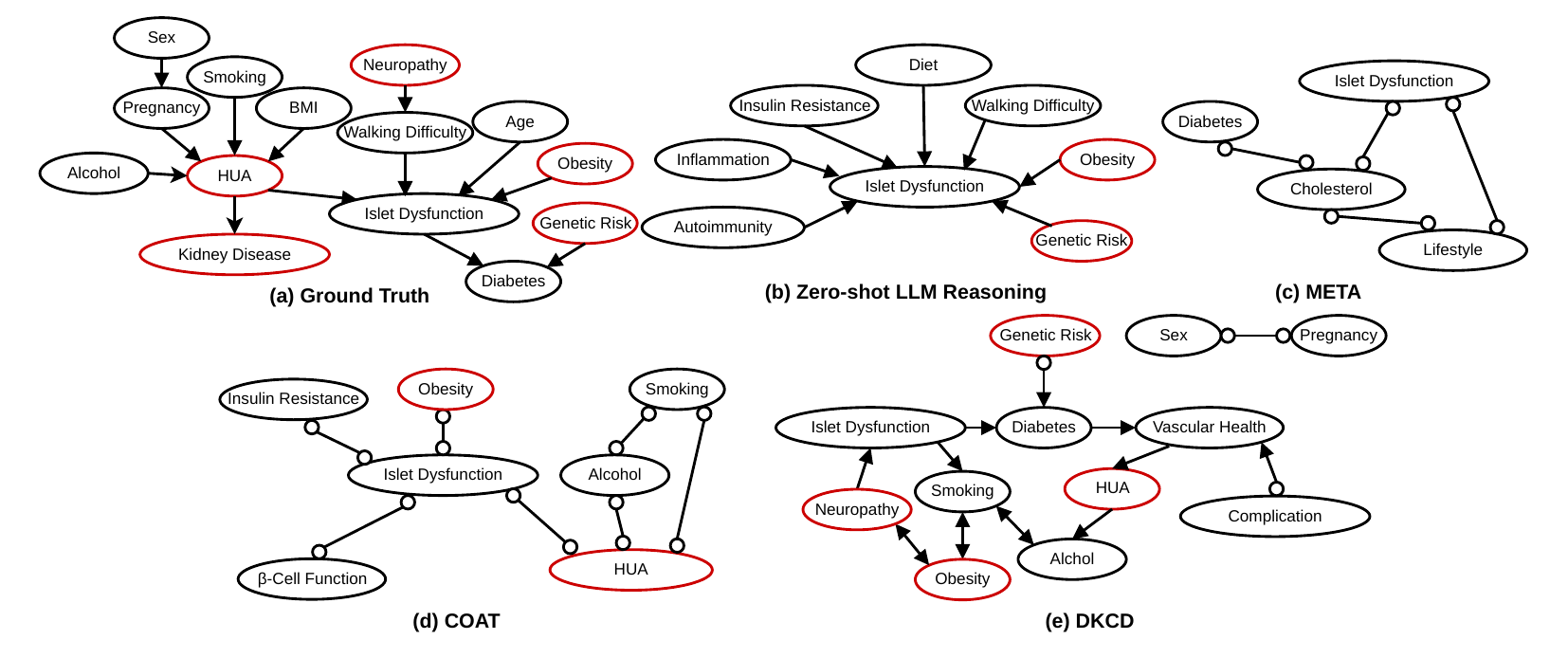}
    \caption{Ground-truth and discovered causal graphs on the diabetes dataset. Latent factors are marked in red.}
    \label{fig:DiabetesCG}
    \vspace{-6pt}
\end{figure*}

The empirical results of causal discovery on the diabetes and respiratory datasets are presented in Table~\ref{tab:DiaResults} and Figure~\ref{fig:DiabetesCG}, and Table~\ref{tab:ResResults} and Figure~\ref{fig:RespiratoryCG_GPT}, respectively. The results demonstrate that DKCD consistently outperforms the baseline methods in most cases while maintaining stable performance across different LLM backbones. 

\noindent
\textbf{Results on the Diabetes Dataset:} DKCD achieves better overall causal factor identification performance across most LLM settings, obtaining higher NP, NR, and NF scores than the baselines, as shown in Table~\ref{tab:DiaResults}. Compared with COAT \cite{liu2024discovery}, DKCD more accurately identifies both observable and latent causal factors, showing the effectiveness of incorporating domain KGs and latent factor reasoning for causal discovery from unstructured data. DKCD also achieves competitive graph structure performance, with lower ESHD values and stable AR and AF scores across different LLM backbones, indicating closer alignment with the ground-truth causal graph, as illustrated in Figure~\ref{fig:DiabetesCG}. Additional graph comparison results across different LLM backbones are provided in Appendix~\ref{appendix:Full_results}.


\noindent
\textbf{Results on the Respiratory Dataset:} A similar trend can be observed on the respiratory dataset. In terms of causal factor identification, DKCD achieves the highest NP, NR, and NF scores across most LLM configurations, with quantitative results presented in Table~\ref{tab:ResResults}. These results indicate that DKCD has a strong ability to extract meaningful causal factors from patient descriptions of respiratory conditions. In terms of causal graph structure generation, DKCD also achieves lower ESHD values and higher AR and AF scores in most cases. These results indicate that DKCD can effectively recover the underlying causal structure among respiratory-related factors. Overall, the results further confirm that integrating domain knowledge with causal discovery algorithms significantly improves causal discovery from unstructured data. Additional graph comparison results across different LLM backbones are provided in Appendix~\ref{appendix:Full_results}.


\subsection{Ablation Study}
We conduct an ablation study using GPT-4o as the representative LLM backbone to evaluate the effectiveness of key components in DKCD. Specifically, we compare the full DKCD model with three variants: (1) w/o CC removes causal clues in the knowledge-guided causal reasoning module; (2) w/o LF removes latent factor support, causing the annotation process to rely solely on LLM-identified factors; and (3) w/o Both removes both causal clues and latent factor identification, simulating the setting without knowledge graphs. Additional ablation results and analyses of different design choices are provided in Appendix~\ref{appendix:ablation_results}.

The experimental results on the Diabetes and Respiratory datasets are shown in Table~\ref{tab:AblationResults}. Removing causal clues mainly degrades structural metrics such as AF and ESHD, while the NF score remains unchanged because causal clues are introduced after factor mining and therefore do not affect node identification. In contrast, removing latent factor identification leads to larger performance drops, especially in NF, highlighting the importance of domain knowledge for discovering causal latent factors. Removing both components causes the severe degradation, indicating that the two designs are complementary. Overall, DKCD achieves competitive performance across both datasets, demonstrating the benefit of integrating domain knowledge with reasoning guidance for causal discovery.

\begin{table}[t]
\centering
\small
\setlength{\tabcolsep}{3pt}
\caption{Ablation study of the DKCD framework on the Diabetes and Respiratory datasets.}
\label{tab:AblationResults}
\resizebox{\columnwidth}{!}{%
\begin{tabular}{llccc}
\toprule
Dataset & Variant & NF$\uparrow$ & AF$\uparrow$ & ESHD$\downarrow$ \\
\midrule
\multirow{4}{*}{Diabetes dataset}
& w/o CC  & \textbf{0.77 $\pm$ 0.04} & 0.38 $\pm$ 0.15 & 26.67 $\pm$ 2.52 \\
& w/o LF   & 0.33 $\pm$ 0.14 & 0.00 $\pm$ 0.00 & 26.67 $\pm$ 2.52 \\
& w/o Both & 0.54 $\pm$ 0.31 & 0.10 $\pm$ 0.09 & 34.00 $\pm$ 1.73 \\
& DKCD  & \textbf{0.77 $\pm$ 0.04} & \textbf{0.40 $\pm$ 0.18} & \textbf{25.33 $\pm$ 3.21} \\
\midrule
\multirow{4}{*}{Respiratory dataset}
& w/o CC  & \textbf{0.74 $\pm$ 0.07} & 0.62 $\pm$ 0.10 & 17.00 $\pm$ 2.65 \\
& w/o LF   & 0.45 $\pm$ 0.11 & 0.32 $\pm$ 0.16 & 22.33 $\pm$ 4.51 \\
& w/o Both & 0.33 $\pm$ 0.15 & 0.14 $\pm$ 0.12 & 25.33 $\pm$ 1.53 \\
& DKCD  & \textbf{0.74 $\pm$ 0.07} & \textbf{0.68 $\pm$ 0.21} & \textbf{16.67 $\pm$ 3.21} \\
\bottomrule
\end{tabular}%
}
\vspace{-10pt}
\end{table}

\section{Conclusions}
This paper presents DKCD, a knowledge-enhanced framework for causal discovery from unstructured data in high-expertise domains. DKCD consists of three key components: a knowledge mining module that retrieves relevant domain knowledge, a knowledge-guided causal reasoning module that discovers latent causal factors and generates key causal clues for accurate annotation, and a causal structure discovery module that constructs the final causal graph from the annotation table. Extensive experiments on two domain datasets show that DKCD significantly outperforms existing methods.




\section*{Limitations}
Although DKCD integrates LLMs with domain knowledge graphs to improve causal discovery from unstructured data, several limitations remain.

\noindent
\textbf{Reliance on LLM-generated reasoning.}
DKCD still relies on LLMs' reasoning and generation capabilities. Although our design provides domain-knowledge-grounded reasoning, LLMs may still generate hallucinated, inconsistent, or biased causal factors and causal clues, especially in complex domain-specific scenarios.

\noindent
\textbf{Limited availability of benchmark datasets.}
Causal discovery from unstructured domain-specific data is limited by the lack of publicly available benchmark datasets containing both unstructured textual descriptions and aligned ground-truth causal graphs. Constructing reliable benchmarks in specific domains remains challenging because it requires extensive domain expertise, causal annotation, and validation \cite{visbal2004gender, rodrigues2022reflection}. Although the datasets used in this work are carefully constructed based on prior literature and grounded in real-world medical knowledge, they should be viewed as a starting point for developing and evaluating methods in this emerging direction of domain-specific unstructured causal discovery. Developing larger-scale and more diverse benchmarks will be a focus of future work.

\noindent
\textbf{Scalability to broader domains.}
Our experiments focus on two medical-domain datasets, which allow us to evaluate DKCD in specialized settings where domain knowledge is important. However, the generalization of DKCD to other domains, such as education, finance, and the social sciences, remains underexplored.

\noindent
\textbf{Future work} will address these limitations in three main directions. First, we will develop stronger validation mechanisms for LLM-generated causal reasoning, including consistency checking, uncertainty estimation, and causal constraints derived from domain knowledge. Second, we will expand the evaluation of DKCD by constructing and testing on more diverse datasets, including real-world domain-specific textual data, larger-scale knowledge graphs, and expert-validated causal structures. Third, we will investigate the application of DKCD to broader specific domains.

\section*{Ethical Considerations}
This research focuses on methodological work for constructing causal graphs from unstructured textual data. The datasets used in our experiments are either synthetic or derived from publicly available sources, and do not contain personally identifiable information. Additionally, our framework is intended to assist researchers in discovering potential causal structure and should not be used as a substitute for professional judgment, particularly in sensitive domains such as healthcare. Future applications should carefully consider issues related to data privacy, the potential biases inherent in language models, and the responsible generation, interpretation, and use of automatically inferred causal relationships in practical settings.




\bibliography{custom}

\appendix

\section{Algorithm}
\label{section:algorithm}

\begin{center}
\begin{minipage}{\columnwidth}
\hrule
\vspace{0.3em}
\textbf{Algorithm 1:} The DKCD Framework
\vspace{0.3em}
\hrule
\vspace{0.5em}

\textbf{Input:} $\mathcal{D}$, $\mathcal{X}_s$, $\mathcal{G}_{KG}$, $f_{\mathrm{obs}}$, $M$, $L$, $\mathcal{C}$.

\textbf{Output:} $\mathcal{V}$, $\mathbf{S}$, $G$.

\begin{enumerate}[leftmargin=*,itemsep=0.2em,topsep=0.2em]

\item Propose observable factors:
$\mathcal{V}_o \leftarrow f_{\mathrm{obs}}(\mathcal{X}_s)$.

\item Initialize
$\mathbf{S} \leftarrow [\ ]$.

\item For each $\mathbf{x}_k \in \mathcal{D}$:
\begin{enumerate}[label=(\alph*),leftmargin=*,itemsep=0.15em,topsep=0.15em]

\item Retrieve subgraph:
$\mathcal{G}_{rel}^{(k)} \subseteq \mathcal{G}_{KG}$.

\item Verbalize subgraph:
$\mathcal{S}^{(k)}=\{s_1,\dots,s_{n_k}\}$.

\item Semantic matching:
$\mathcal{S}_{v_i}^{(k)} \leftarrow M(v_i,\mathcal{S}^{(k)})$,
$\forall v_i \in \mathcal{V}_o$.

\item Build matched context:
$\widetilde{\mathcal{S}}^{(k)}
=
\{\mathcal{S}_{v_i}^{(k)} \mid v_i \in \mathcal{V}_o\}$.

\item Discover latent factors:
$\mathcal{V}_l
\leftarrow
L(\mathcal{V}_o,\widetilde{\mathcal{S}}^{(k)})$.

\item Generate causal clues:
$\mathcal{H}
\leftarrow
L(\mathcal{V}_o \cup \mathcal{V}_l,\mathcal{G}_{rel}^{(k)})$.

\item Annotate factors:
$\mathcal{V}
\leftarrow
\mathcal{V}_o \cup \mathcal{V}_l$,

$\mathbf{s}^{(k)}
\leftarrow
(v_{k1},\dots,v_{k|\mathcal{V}|})$.

\item Update scoring table:
$\mathbf{S}
\leftarrow
\mathbf{S}
\cup
\{\mathbf{s}^{(k)}\}$.

\end{enumerate}

\item Discover causal graph:
$G \leftarrow \mathcal{C}(\mathbf{S},\mathcal{V})$.

\item Return $\mathcal{V}, \mathbf{S}, G$.

\end{enumerate}

\vspace{0.3em}
\hrule
\end{minipage}
\end{center}

\section{More Details about Experiments}
\label{sec:more_experiment_details}

We construct two medical-domain datasets: the Diabetic Patient Condition Descriptions Dataset (Diabetes Dataset) and the Respiratory Patient Condition Descriptions Dataset (Respiratory Dataset). Each dataset contains 400 patient condition descriptions, a set of high-level factors, and a corresponding domain knowledge graph.

\subsection{Dataset Construction Process}
\label{appendix:dataset_examples}

\subsubsection{Diabetes Dataset}
\label{appendix:diabetesConstruct}

\begin{figure*}[t]
    \centering
    \includegraphics[width=\textwidth]{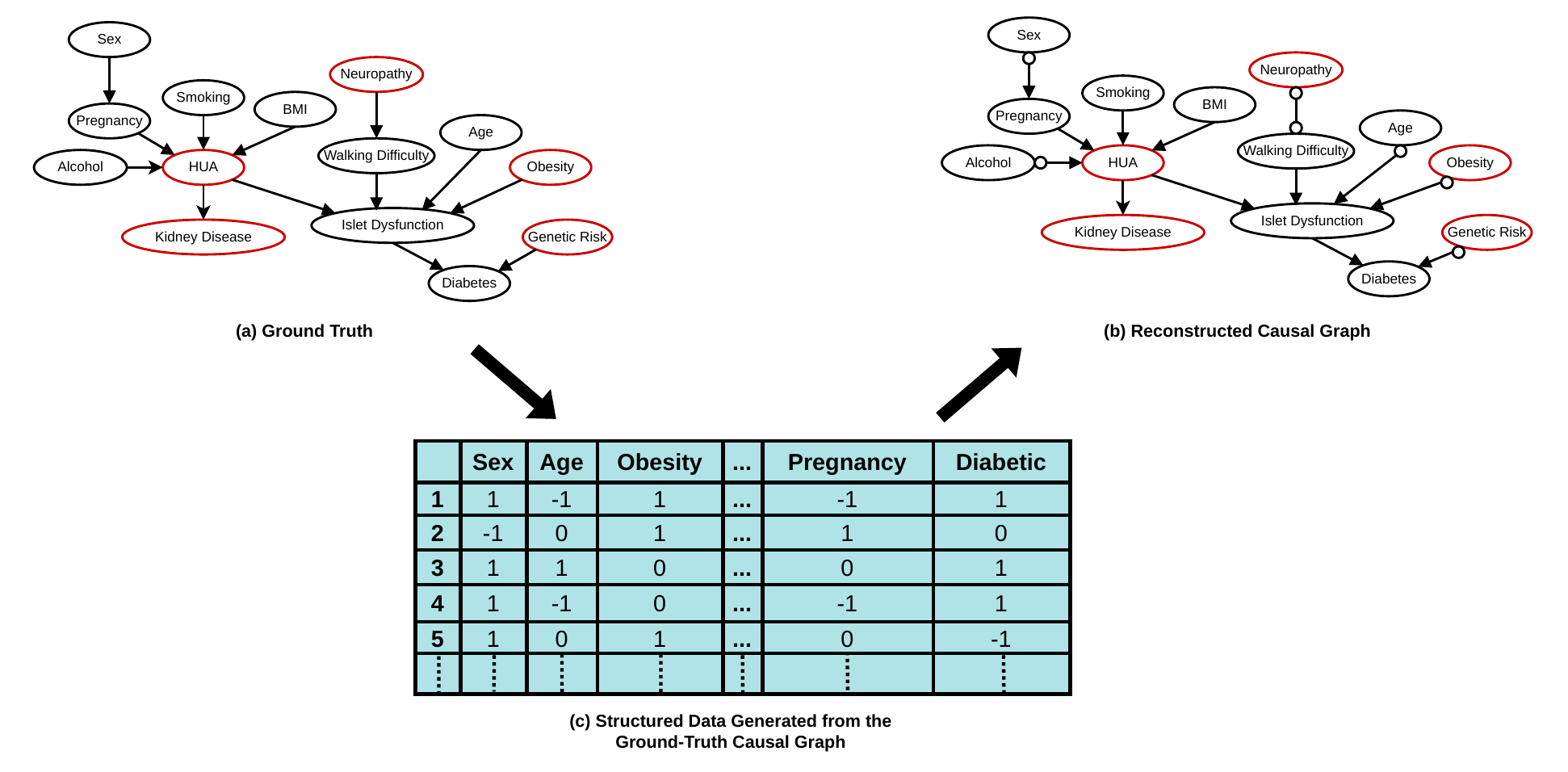}
    \caption{Ground truth causal graph and the causal graph reconstructed by the FCI algorithm from the generated structured data derived from the ground truth on the diabetes dataset. Latent factors are marked in red.}
    \label{fig:DiabetesGenerate}
\end{figure*}

\noindent
\textbf{Diabetic Patient Condition Descriptions Generation:} 
We construct a dataset of diabetic patient condition descriptions following the methodology proposed in \cite{liu2024discovery}. As illustrated in Figure~\ref{fig:DiabetesGenerate}, we first construct a ground-truth causal graph based on domain-specific academic materials \cite{noh2025diabetes, molina2021deciphering, meijnikman2018evaluating}, and then generate structured data samples in the form of a scoring table according to the predefined causal relationships encoded in the graph. The FCI algorithm is then applied to the generated structured dataset to reconstruct the causal graph for causal consistency verification. Only when the reconstructed graph is structurally consistent with the ground-truth causal graph are the structured samples used to generate descriptions of diabetic patients' conditions. Finally, Prompt~\ref{prompt:diabetes} is used to convert the generated structured samples into unstructured diabetic patient condition descriptions.

\noindent
\textbf{Example Generated Diabetic Patient Condition Description:} 
\label{appendix:diabetesExample} 
An example generated diabetic patient condition description is shown below:

\begin{tcolorbox}[breakable, enhanced, colback=gray!15, colframe=black, boxrule=0.5pt]
A middle-aged woman, a non-smoker with no alcohol use history, presents for routine assessment amidst long-standing diabetes and hypertension. She maintains a BMI of 24 kg/m$^2$, though she experiences activity-related limitations attributed to her weight, potentially suggesting underlying obesity. There is no report of numbness or tingling, inconsistent with neuropathy. Family history reveals no genetic predisposition to diabetes or associated disorders. Vitals: blood pressure is well-controlled at 128/82 mmHg. Laboratory results indicate a fasting plasma glucose of 5.2 mmol/L, 2-hour OGTT of 6.4 mmol/L, and HbA1c at 5.5\%, suggesting normal glucose metabolism. Kidney function tests show an eGFR of 58 mL/min/1.73 m$^2$, with albuminuria at 32 mg/g, suggesting chronic kidney disease. Lipid panel results are within normal limits, showing LDL-C at 2.2 mmol/L, HDL-C at 1.4 mmol/L, and triglycerides at 1.0 mmol/L. Pathophysiologically, derived indices such as HOMA-IR $\approx 1.5$ and HOMA-$\beta$ $\approx 80\%$ indicate maintained insulin sensitivity and robust $\beta$-cell function. Cardiometabolic risk context is moderate, influenced by her weight considerations and renal findings but offset by optimal lipid levels and normoglycemic status. Diabetes status: absent; metrics do not signify diabetes, sustaining a diagnosis of normoglycemia.
\end{tcolorbox}

\noindent
\textbf{Diabetes-Domain KG Construction:} 
We construct the diabetes-domain KG using Neo4j \cite{miller2013graph}, a graph database technology, based on the diabetes dataset provided in \cite{chang2021diakg}, , which contains annotated entities and relations for diabetes-related medical knowledge.

\subsubsection{Respiratory Dataset}
\label{appendix:RespiratoryConstruct}

\begin{figure*}[t]
    \centering
    \includegraphics[width=\textwidth]{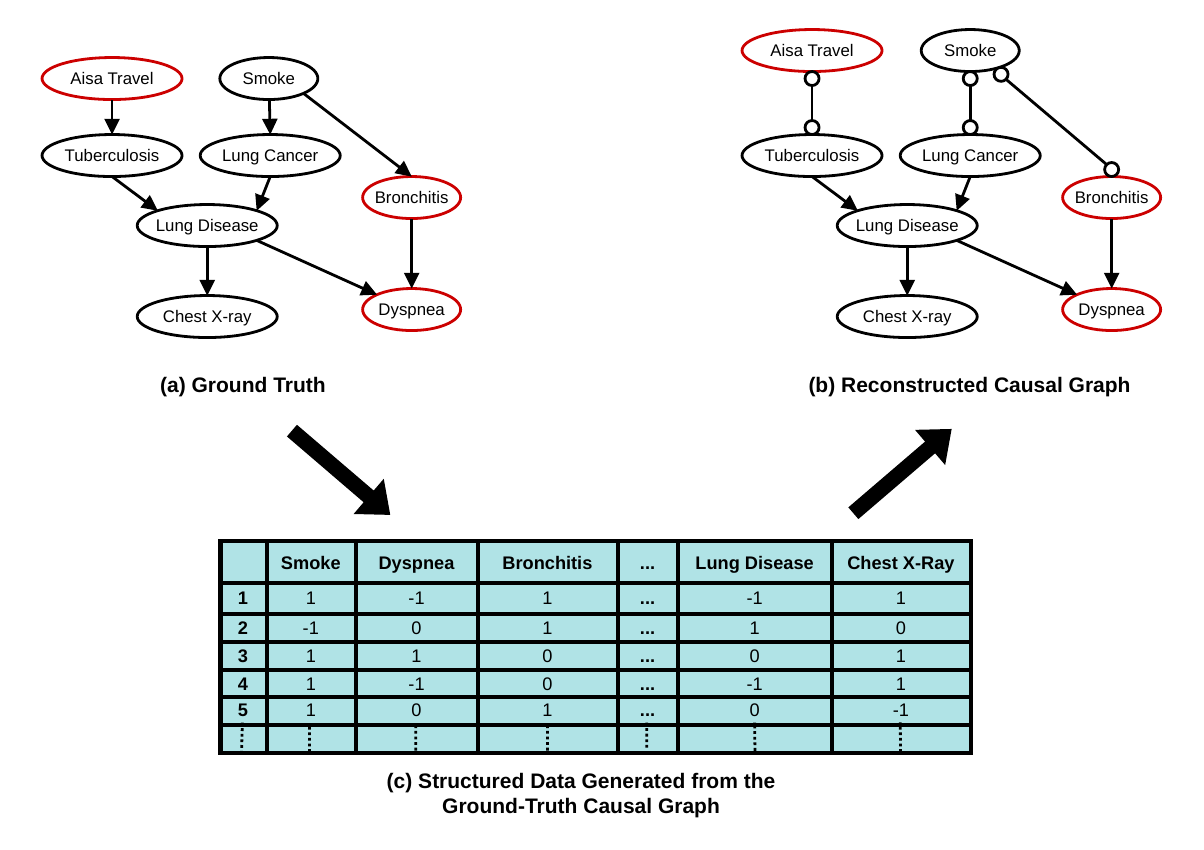}
    \caption{Ground-truth causal graph, generated structured dataset, and the causal graph reconstructed by the FCI algorithm on the respiratory dataset. The ground-truth graph is first used to generate structured data samples, from which the FCI algorithm reconstructs the causal graph. Latent factors are highlighted in red.}
    \label{fig:RespiratoryGenerate}
\end{figure*}

\noindent
\textbf{Respiratory Patient Condition Descriptions Generation:} 
We construct a dataset of respiratory patient condition descriptions following the methodology proposed in \cite{liu2024discovery}. As illustrated in Figure~\ref{fig:RespiratoryGenerate}, we first construct a ground-truth causal graph based on prior domain knowledge and the causal structure introduced in \cite{lauritzen1988local}, and then generate structured data samples in the form of a scoring table according to the predefined causal relationships encoded in the graph. The FCI algorithm is then applied to the generated structured dataset to reconstruct the causal graph for verification of causal consistency. Only when the reconstructed graph is structurally consistent with the ground-truth causal graph are the structured samples used to generate corresponding descriptions of respiratory patient conditions. Finally, Prompt~\ref{prompt:respiratory} is used to convert the generated structured samples into unstructured respiratory patient condition descriptions.

\noindent
\textbf{Example Generated Respiratory Patient Condition Description:} 
An example generated respiratory patient condition description is shown below:
\label{appendix:respiratoryExample} 

\begin{tcolorbox}[breakable, enhanced, colback=gray!15, colframe=black, boxrule=0.5pt]
A 34-year-old woman presents to the clinic for a routine check-up without any specific complaints. She denies recent travel, including visits to Asia, or any known exposure to tuberculosis. The patient is a non-smoker and has no significant respiratory history. She reports being in good health overall, with no history of cough, sputum production, fever, weight loss, night sweats, or hemolysis. Additionally, she does not experience dyspnea, either at rest or with exertion, and her daily activities are not limited. Vital signs are stable, with a temperature of 36.7$^\circ$C, heart rate 72 bpm, respiratory rate 16/min, blood pressure 118/76 mmHg, and oxygen saturation 98\% on room air. Chest examination reveals normal breath sounds with no wheeze or crackles. Routine laboratory tests show a normal white blood cell count and inflammatory markers. A chest X-ray performed as part of her health maintenance check reveals normal findings without evidence of infiltrates or abnormal pulmonary opacities. Clinical impression indicates no underlying lung condition, and the chest X-ray supports this normal finding. She is reassured, and no further pulmonary investigation is needed at this time. Recommendations include maintaining a healthy lifestyle and routine follow-up as part of preventive healthcare.
\end{tcolorbox}

\noindent
\textbf{Respiratory-Domain KG Construction:} 
Since suitable respiratory-domain knowledge graphs are difficult to obtain, we collect a set of publications in the respiratory domain \cite{rahman2023association, stocks1959cancer, lauritzen1988local, onozaki2015national, caplin1975relation}. Based on these publications, we employ an LLM (GPT-4o) to extract relational triples in the form of (entity, relation, entity), which serve as the basic structure for KG construction \cite{pujara2013knowledge, zou2020survey}. The extracted triples are then used Neo4j \cite{miller2013graph} graph database technology to construct the respiratory-domain KG.

\subsection{Details on Prompts}
\label{appendix:prompts}

In this section, we provide examples of prompts used in our experiments, including:

\begin{itemize}[leftmargin=1.2em, itemsep=8pt, topsep=8pt, parsep=0pt, partopsep=0pt]
    \item Generating diabetic patient condition descriptions for the diabetes dataset (Prompt~\ref{prompt:diabetes}).
    \item Generating respiratory patient condition descriptions for the respiratory dataset (Prompt~\ref{prompt:respiratory}).
    \item An example of observable causal factor mining (Prompt~\ref{prompt:observable_mining}).
    \item An example of knowledge-guided latent causal factor mining (Prompt~\ref{prompt:latent_mining}).
    \item An example of causal clues generation for causal structure discovery (Prompt~\ref{prompt:causal_clue}).
    \item An example of the text evaluation prompt 
    (Prompt~\ref{prompt:text_eva}).
\end{itemize}

\refstepcounter{prompt}
\begin{promptbox}{Generating Diabetic Patient Condition Descriptions for the Diabetes Dataset}
\label{prompt:diabetes}
\small

\textbf{Task:}

Please write a clinical description for a patient based on the given diabetic evaluation results.

Evaluation Results: \{Diabetic Factors\}

\textbf{Requirement:}

- Integrate all provided factors into a clinically plausible patient condition description.

- Embed natural clinical clues that imply the following latent factors without explicitly labeling them:

    \begin{itemize}
        \item Neuropathy: numbness, tingling, burning sensation, or reduced sensation in the feet.
        \item Obesity: elevated body weight, increased BMI, or weight-related mobility limitations.
        \item HUA (Hyperuricemia): gout attacks, elevated serum urate, or urate-lowering medication.
        \item ChronicKidneyDisease: albuminuria, reduced eGFR, kidney impairment, or long-standing metabolic disease.
        \item RiskGenes: family history of diabetes or genetic predisposition to metabolic disorders.
    \end{itemize}

- Single paragraph; No quotation marks; The description should be complete.

- Modern clinical English.

- 150--250 words.

- Only output the patient condition description directly without any other format or content.
\end{promptbox}

\refstepcounter{prompt}
\begin{promptbox}{Generating Respiratory Patient Condition Descriptions for the Respiratory Dataset}
\label{prompt:respiratory}
\small

\textbf{Task:}

Please write a clinical description for a patient based on the given respiratory evaluation results.

Evaluation Results: \{Respiratory Factors\}

\textbf{Requirement:}

- Integrate all provided factors into a clinically plausible patient condition description.

- Embed natural clinical clues that imply the following latent factors without explicitly labeling them:

    \begin{itemize}
    \item Visit\_to\_Asia: travel or exposure history related to TB risk.
    \item Bronchitis: cough, sputum production, wheezing, or bronchitic symptoms.
    \item Dyspnea: degree of shortness of breath or breathing difficulty.
    \end{itemize}

- Single paragraph; No quotation marks; The description should be complete.

- Modern clinical English.

- 150--250 words.

- Only output the patient condition description directly without any other format or content.
\end{promptbox}

\refstepcounter{prompt}
\begin{promptbox}{An Example of Observable Causal Factor Mining}
\label{prompt:observable_mining}
\small

You are an expert clinical analyst specializing in medical record interpretation. You are ready to analyze contrasting patient condition descriptions and identify observable causal factors.  

\textbf{Task:}

Patient Condition Descriptions: \{Patient Condition Descriptions\}

Based on these patient condition descriptions, analyze the underlying differences among causal factors that may contribute to the observed distinctions.

\textbf{Requirements:}

- Identify the key observable factors that differentiate the patient records.

- Each factor should focus on one concrete aspect without semantic overlap.

- Keep the criterion for value 0 fixed as: not exist; or not mentioned.

- Each factor is only allowed to take one value from \{-1, 0, 1\}.

\textbf{Output Format:}

Your output should contain the following three parts.

\textbf{Part 1: Consideration}

Explain your reasoning process when identifying causal factors from the patient records.  
Focus on how the textual evidence supports the abstraction of candidate factors.

\textbf{Part 2: Factor Filtration}

Evaluate each candidate factor and decide whether to retain it.

Selection causal factor criteria:
\begin{itemize}
\item Each factor should help distinguish different patient conditions.
\item Each factor should represent a single concrete aspect.
\item Semantic overlap between factors should be minimized.
\end{itemize}

\textbf{Part 3: Final Output}

Report the final factors using the following template:

\{Factor Name\}
    \begin{itemize}
    \item - 1: [Positive Criterion]
    \item - 0: [Not exist; or not mentioned]
    \item - -1: [Negative Criterion]
    \end{itemize}

\end{promptbox}

\refstepcounter{prompt}
\begin{promptbox}{An Example of Knowledge-guided Latent Causal Factor Mining}
\label{prompt:latent_mining}
\small

Use the patient condition description, relevant domain knowledge, and observable causal factor criteria to identify clinically meaningful \textbf{latent causal factors}.

Patient Condition Descriptions: \{Patient Condition Descriptions\}

Relevant Domain Knowledge: \{Relevant Domain Knowledge\}

Observable Causal Factor Criteria: \{Observable Causal Factor Criteria\}

\textbf{Task:}

Based on the patient condition description and the relevant domain knowledge, identify \textbf{latent causal factors}. These latent factors should capture deeper causal abstractions rather than directly repeating observable factors.

\textbf{Requirements}
\begin{itemize}
\item Latent factors should represent higher-level mechanisms or conceptual causes.
\item They must be distinct from observable factors.
\item Do not duplicate, paraphrase, or rename observable factors.
\item Avoid semantic overlap with observable factors.
\end{itemize}

\textbf{Output Format:}

\textbf{Part 1: Consideration}

Briefly describe how you abstract latent factors from observable factors and domain knowledge.

\textbf{Part 2: Factor Filtration}

Explain why each candidate factor should be kept or discarded.

\textbf{Part 3: Final Output}

Report the final factors using the following template:

\{Factor Name\}
    \begin{itemize}
    \item - 1: [Positive Criterion]
    \item - 0: [Not exist; or not mentioned]
    \item - -1: [Negative Criterion]
    \end{itemize}

\end{promptbox}

\refstepcounter{prompt}
\begin{promptbox}{An Example of Causal Clues Generation for Causal Structure Discovery}
\label{prompt:causal_clue}
\small
\textbf{Task:}

Using ONLY the factors finalized according to \{Patient Condition Description\} and \{Causal Factor Criteria\}, infer causal clues.

Patient Condition Description: \{Patient Condition Description\}

Causal Factor Criteria: \{Causal Factor Criteria\}

Relevant Domain Knowledge: \{Relevant Domain Knowledge\}

\textbf{Requirements}

- Infer causal clues only when there is clear textual evidence from the patient condition description and causal factor criteria.

- Each causal clue should suggest a potential directional relationship between two factors.

- If the relationship is uncertain or unsupported by the text, omit the edge rather than guessing.

\textbf{Output Format}

\textbf{Part 1: Inferred Causal Clues}

List the inferred causal clues in the following format: Source\_Factor → Target\_Factor

\textbf{Part 2: Justification}

Provide a brief justification for each clue based on the patient's condition description.

\end{promptbox}

\refstepcounter{prompt}
\begin{promptbox}{An Example of Text Annotation Prompt}
\label{prompt:text_eva}
\small
\refstepcounter{prompt}
\small
You are a highly capable assistant for analyzing, abstracting, and processing textual data.

\textbf{Task:}

Evaluate the following factors based on the provided patient condition description.

Patient Condition Descriptions: \{Patient Condition Descriptions\}

Each factor must be assigned exactly one value from \{-1, 0, 1\} according to its evaluation criteria.

Factor Criteria and Causal Clue (used as reference during scoring):

\{Factor Criteria\}
\{Causal Clue\}

\textbf{Requirements}

- For each factor, select exactly one value from \{-1, 0, 1\}.

- If the factor is not mentioned in the text, return 0.

- The assigned value must strictly follow the given factor criteria.

\textbf{Output Format}

Return ONLY a single-line compact JSON object:

Rules:
    \begin{itemize}
    \item Output JSON only (no explanations, no markdown, no extra text).
    \item Keys must match exactly (including spaces and lowercase).
    \item Values must be integers from \{-1, 0, 1\}.
    \end{itemize}
    
Example format: \{json\_example\}

\end{promptbox}

\subsection{Implementation Details}
\label{appendix:Implementation}
We implement DKCD as follows. In the knowledge mining module, we first sample representative patient condition descriptions from different outcome groups. Following the COAT method, we empirically select up to 20 samples per group to form a diverse, balanced input set for identifying observable factors. For relevant domain knowledge retrieval, we query the Neo4j domain knowledge database with the extracted observable factors to obtain candidate subgraphs, and then apply semantic matching using the pretrained sentence-transformer model all-MiniLM-L6-v2 \cite{yin2024study, vergou2023readability} to rank retrieved knowledge sentences and retain the top-$r$ most relevant contexts for each factor. Based on the parameter analysis in Table~\ref{tab:TopRStudy}, we set $r=20$ to balance relevant knowledge coverage and contextual noise. In the knowledge-guided causal reasoning module, the LLM takes representative patient condition descriptions, observable factor criteria, and retrieved knowledge contexts as input to infer latent causal factors and causal clues, and further annotates each description in the scoring dataset with a structured scoring vector in $\{-1,0,1\}$. In the causal structure discovery module, we apply the FCI algorithm implemented in the causal-learn library \cite{zheng2024causal} to the annotated scoring table to infer the final causal graph structure. All experimental results are reported as the mean and standard deviation over three independent runs. All experiments are conducted on a server with two Intel Xeon 6346 CPUs, 256GB RAM, and two NVIDIA A40 GPUs, with all LLMs accessed via API calls.

\subsection{Evaluation Metrics}
\label{appendix:metrics}
To evaluate the effectiveness of our method, we adopt a set of commonly used metrics for both causal factor identification and causal graph structure discovery, following prior work \cite{li2025revealing, liu2024discovery, zanga2022survey}.

\noindent
\textbf{Factor Identification:} We evaluate the accuracy and completeness of the identified causal factors using precision (NP), recall (NR), and F1 score (NF). Since factor names may vary across different methods and models, we manually align semantically equivalent factors with the ground-truth names during evaluation to ensure consistency.

\noindent
\textbf{Causal Graph Structure Discovery:} The quality of the discovered causal structure is evaluated using adjacency precision (AP), adjacency recall (AR), and their F1 score (AF). We further adopt the extended structural Hamming distance (ESHD) metric \cite{li2025revealing}. Compared with the standard SHD metric \cite{tsamardinos2006max, norouzi2012hamming}, ESHD additionally accounts for missing and spurious factors and their associated edges.

\subsection{Baselines}
\label{appendix:baselines}

We provide a detailed list of the representative baselines used in our experiments, including:

\begin{itemize}[leftmargin=*, itemsep=8pt, topsep=8pt, parsep=0pt, partopsep=0pt]

\item \textbf{Zero-shot LLM} \cite{antonucci2023zero, du2025causal} relies solely on the reasoning capability of LLMs without providing any examples or domain knowledge.

\item \textbf{META} \cite{liu2024discovery} is a variant derived from the COAT ablation study, where contextual examples are not provided. It performs zero-shot causal factor proposal based only on the given context using LLMs.

\item \textbf{COAT} \cite{liu2024discovery} is a state-of-the-art LLM-driven framework for causal discovery from unstructured text. It integrates LLM reasoning with traditional causal discovery algorithms to construct causal graphs.

\end{itemize}

\subsection{Full Experimental Results}
\label{appendix:Full_results}
\begin{figure*}[t]
    \centering
    \includegraphics[width=\linewidth]{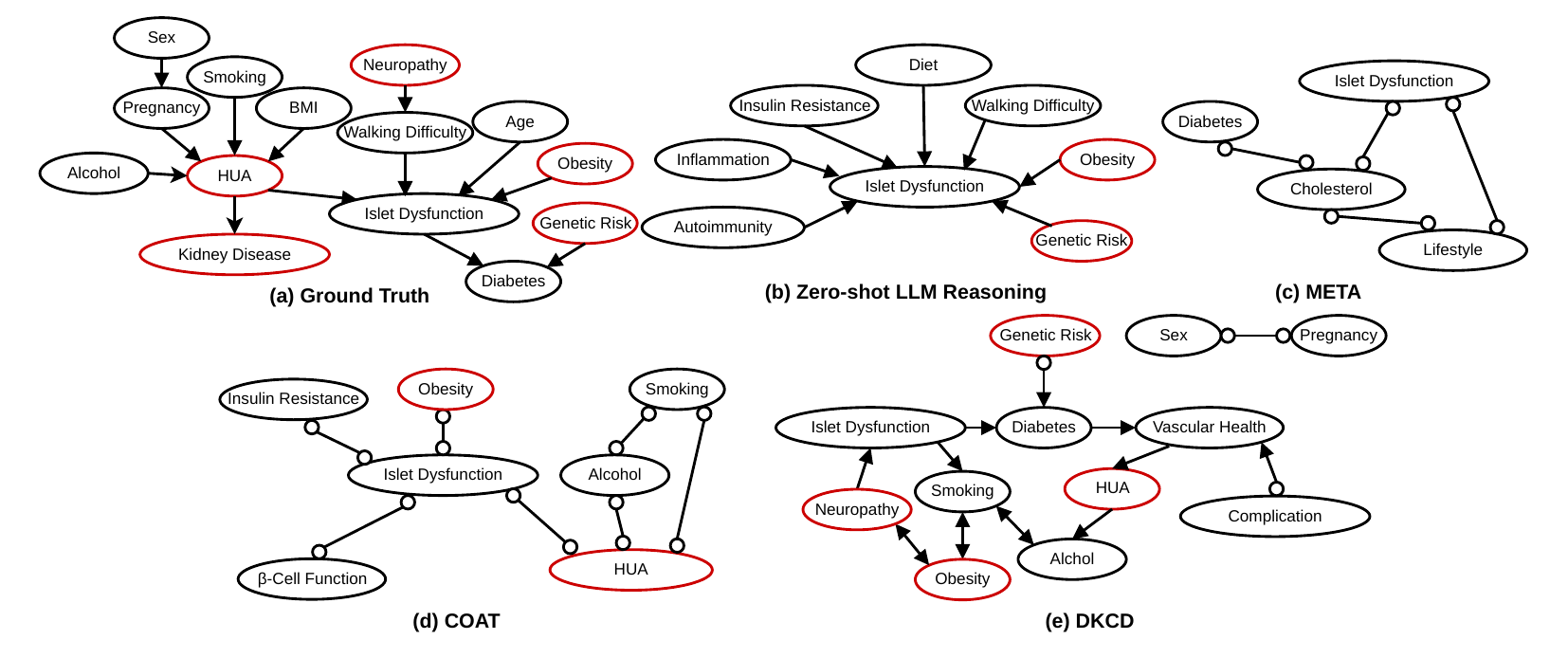}
    \caption{Causal graphs discovered with GPT-4o on the Diabetes dataset. Latent factors are marked in red.}
    \label{fig:DiabetesCG_GPT}
\end{figure*}

\begin{figure*}[t]
    \centering
    \includegraphics[width=\linewidth]{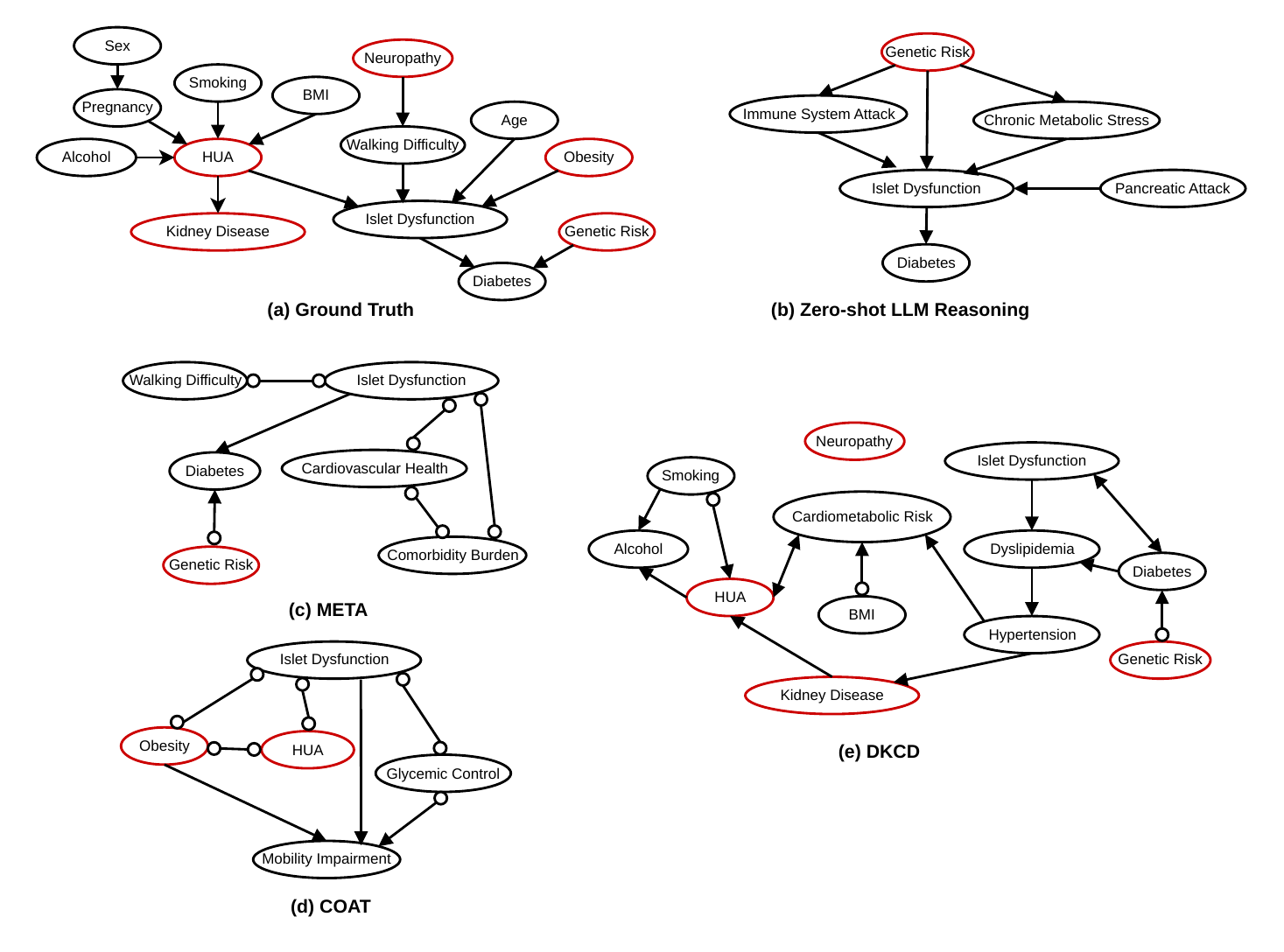}
    \caption{Causal graphs discovered with Gemini 2.5 Flash on the Diabetes dataset. Latent factors are marked in red.}
    \label{fig:DiabetesCG_Gemini}
\end{figure*}

\begin{figure*}[t]
    \centering
    \includegraphics[width=\linewidth]{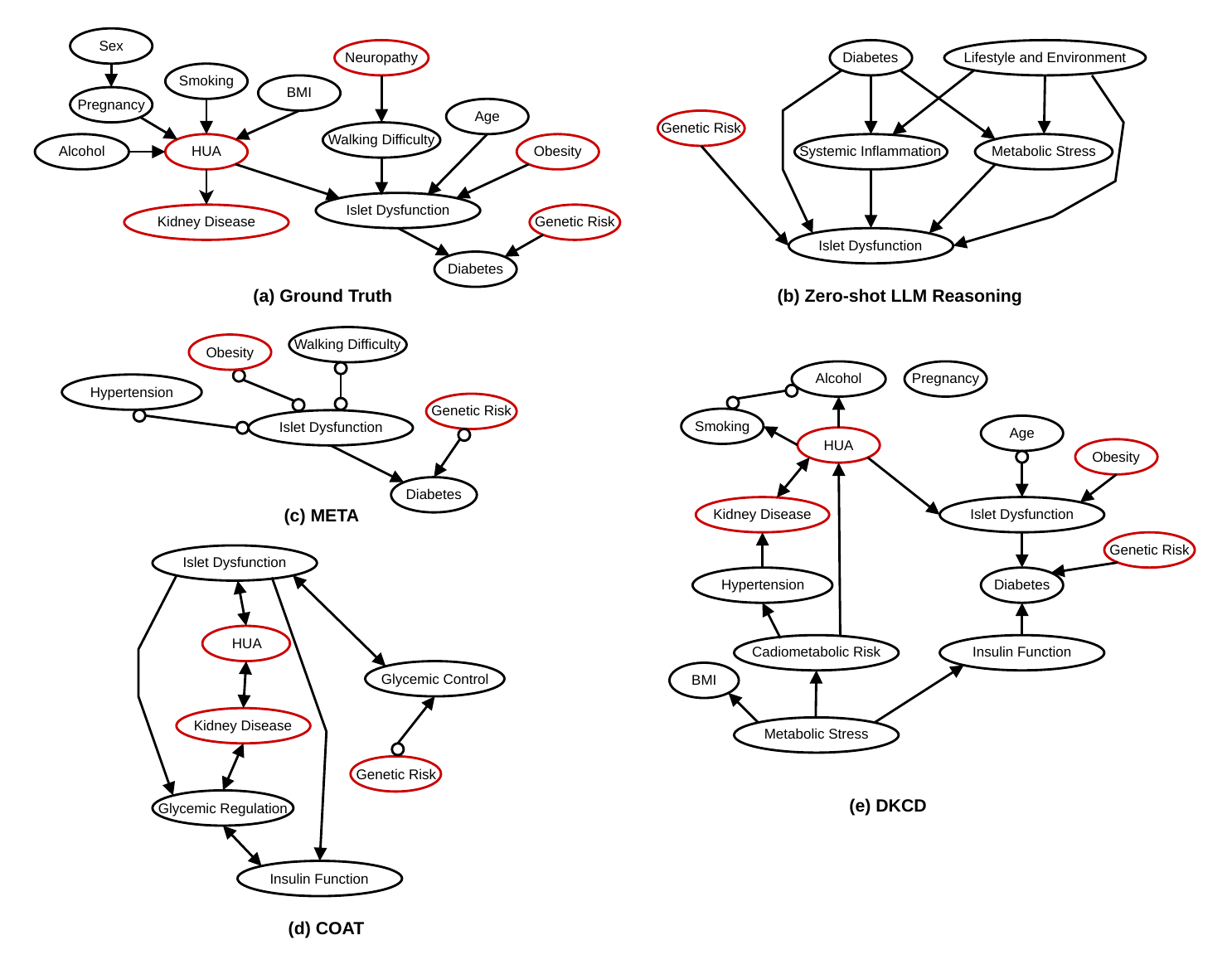}
    \caption{Causal graphs discovered with Grok-3 on the Diabetes dataset. Latent factors are marked in red.}
    \label{fig:DiabetesCG_Grok-3}
\end{figure*}

\begin{figure*}[t]
    \centering
    \includegraphics[width=\linewidth]{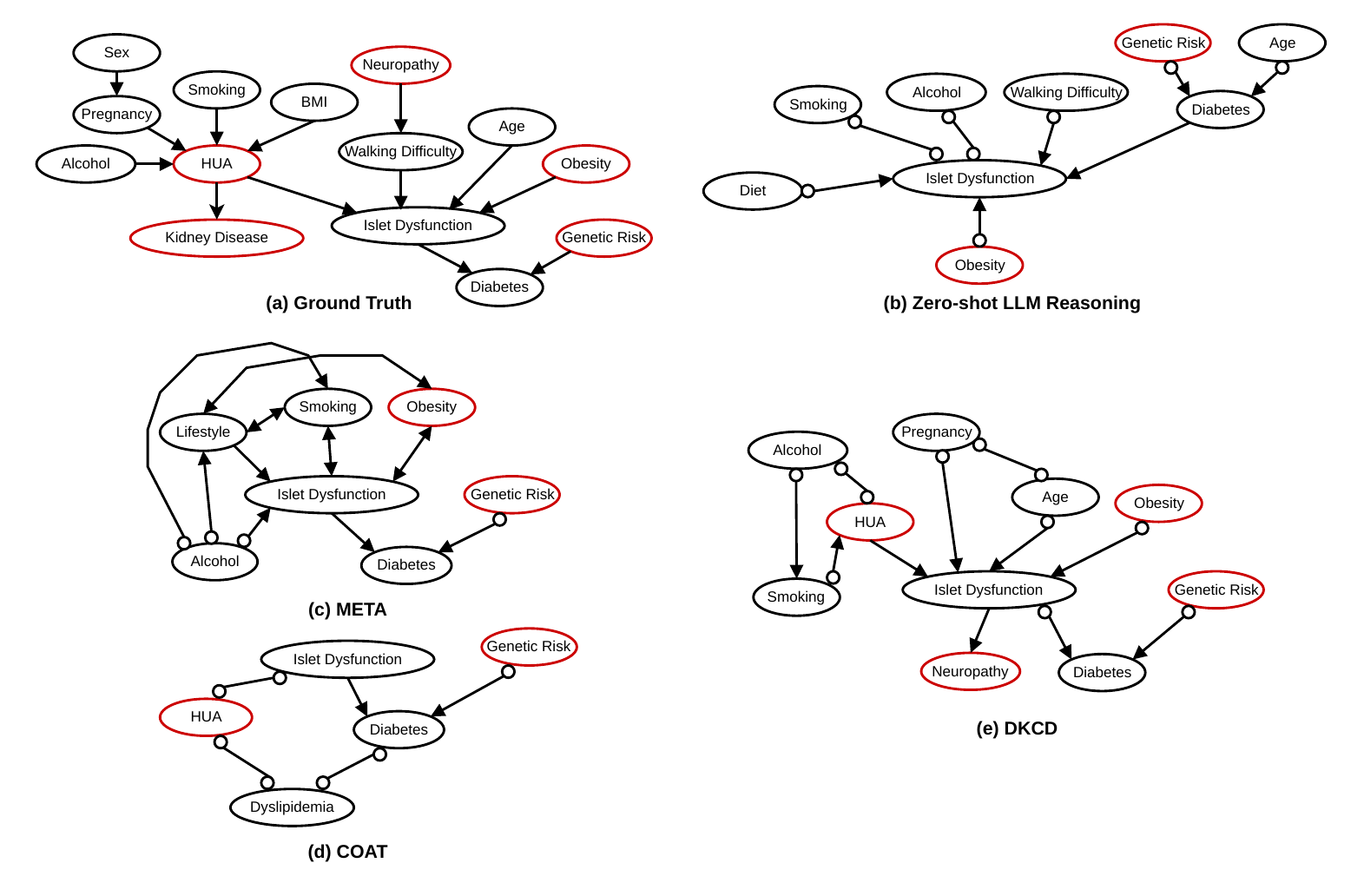}
    \caption{Causal graphs discovered with LLaMA 3-70B on the Diabetes dataset. Latent factors are marked in red.}
    \label{fig:DiabetesCG_LLaMa3}
\end{figure*}

\begin{figure*}[t]
    \centering
    \includegraphics[width=\linewidth]{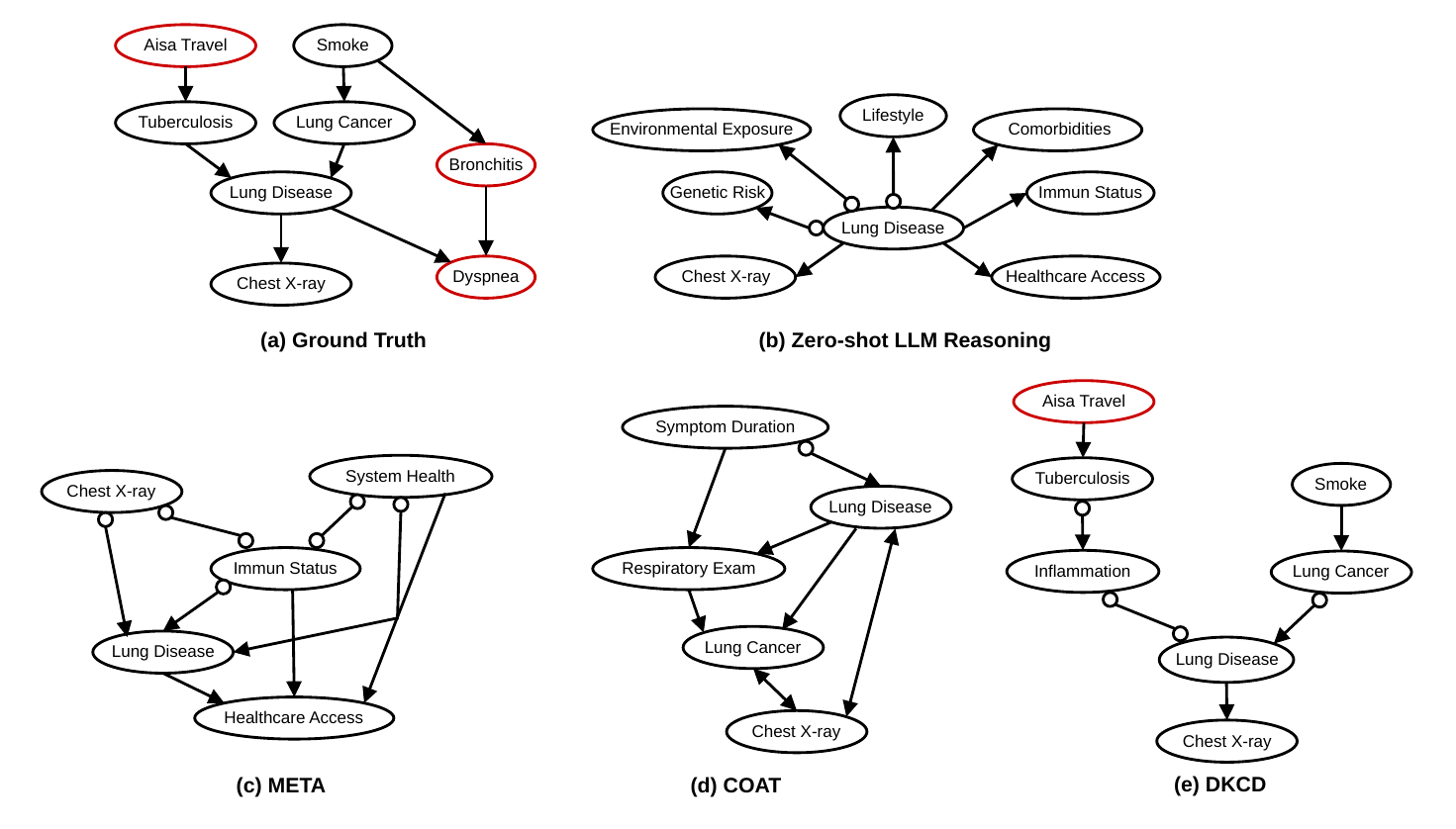}
    \caption{Causal graphs discovered with GPT-4o on the Respiratory dataset. Latent factors are marked in red.}
    \label{fig:RespiratoryCG_GPT}
\end{figure*}

\begin{figure*}[t]
    \centering
    \includegraphics[width=\linewidth]{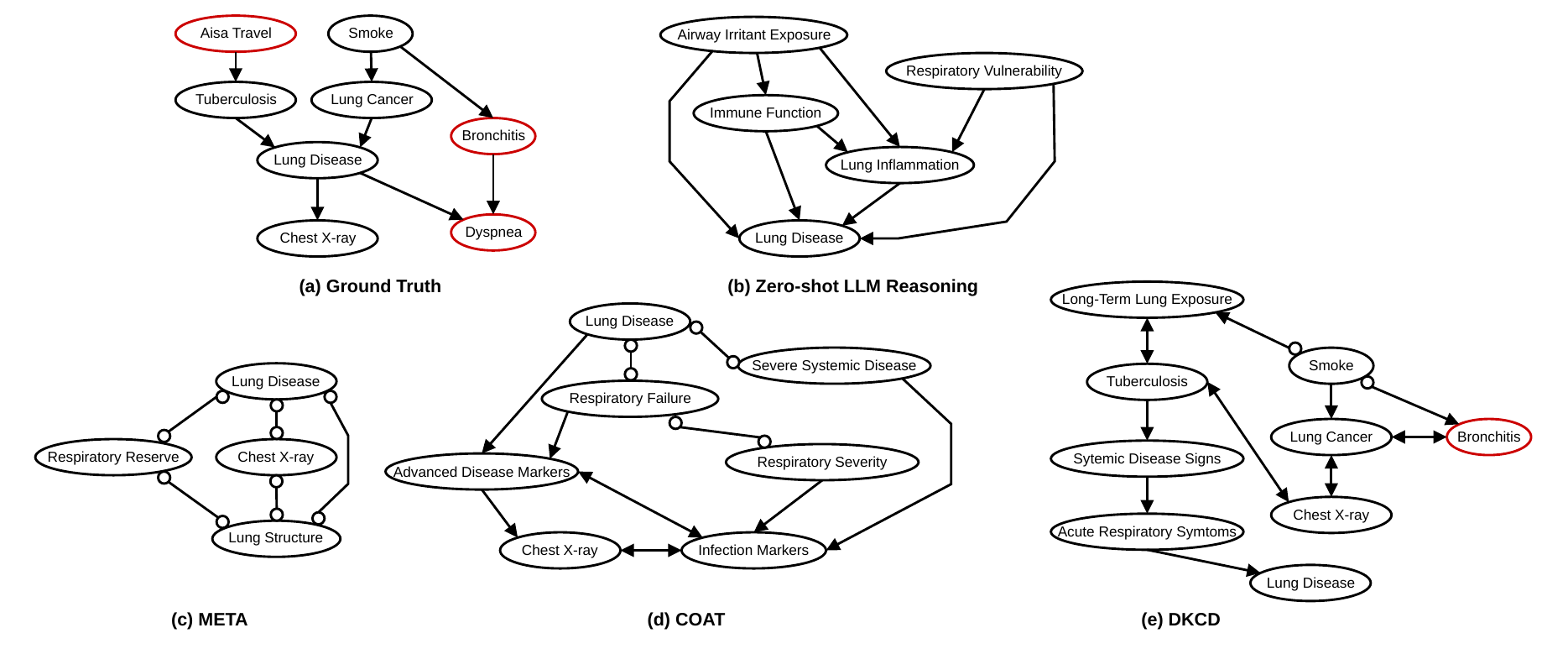}
    \caption{Causal graphs discovered with Gemini 2.5 Flash on the Respiratory dataset. Latent factors are marked in red.}
    \label{fig:RespiratoryCG_Gemini}
\end{figure*}

\begin{figure*}[t]
    \centering
    \includegraphics[width=\linewidth]{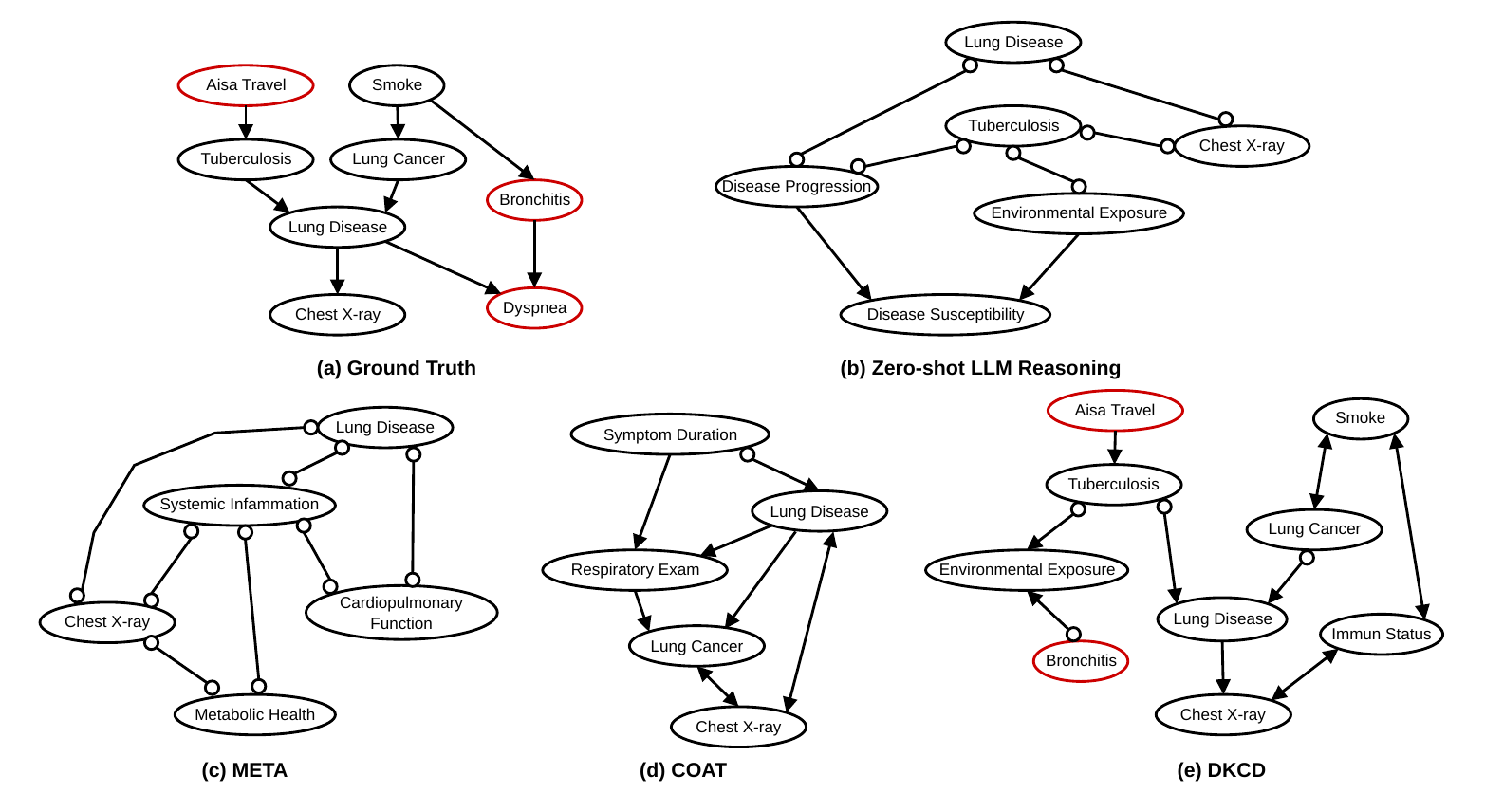}
    \caption{Causal graphs discovered with Grok-3 on the Respiratory dataset. Latent factors are marked in red.}
    \label{fig:RespiratoryCG_Grok-3}
\end{figure*}

\begin{figure*}[t]
    \centering
    \includegraphics[width=\linewidth]{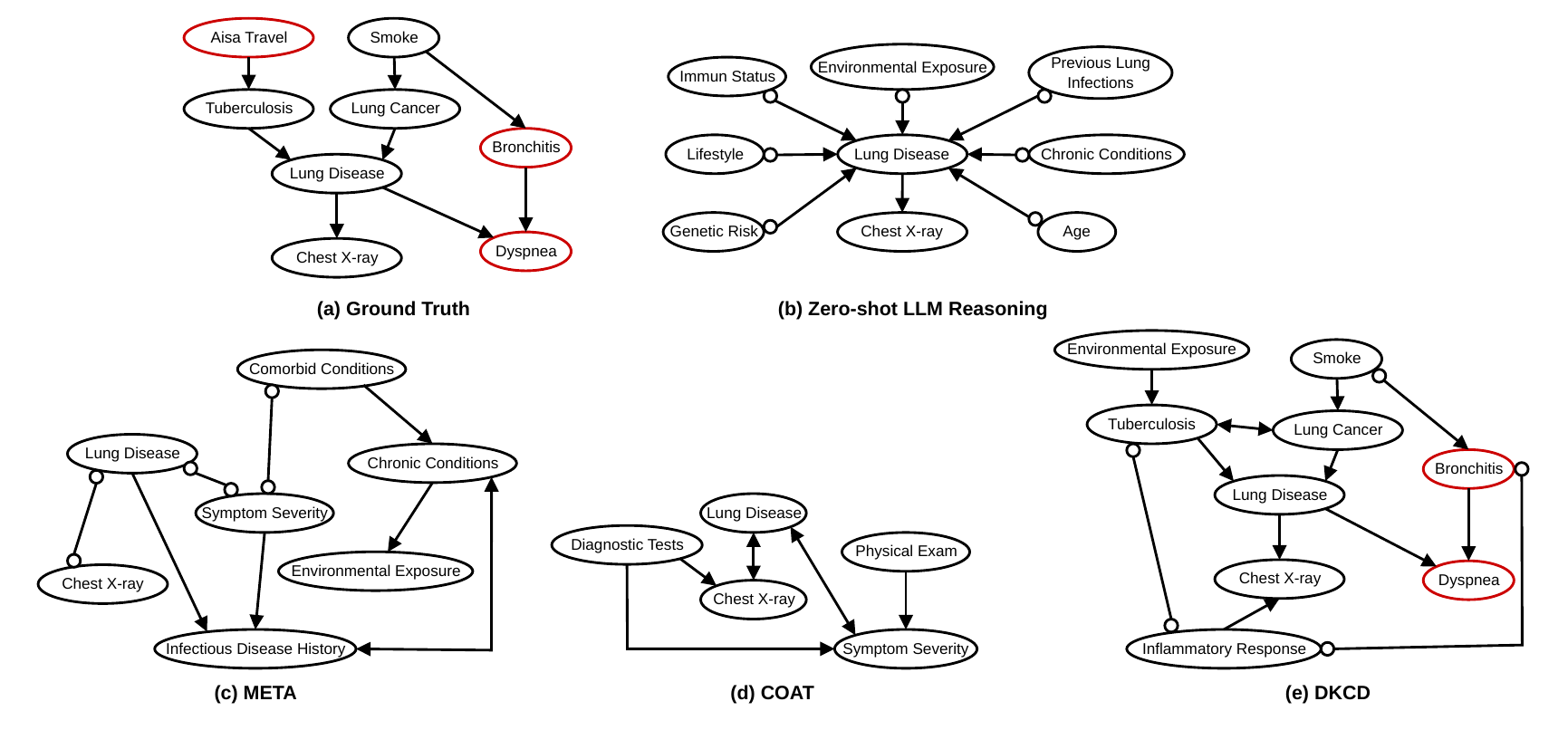}
    \caption{Causal graphs discovered with LLaMA 3-70B on the Respiratory dataset. Latent factors are marked in red.}
    \label{fig:RespiratoryCG_LLaMa3}
\end{figure*}

The complete qualitative comparison of different methods across multiple LLM backbones, including GPT-4o, Gemini 2.5 Flash, Grok-3, and LLaMA 3-70B, on the Diabetes dataset is presented in Figures~\ref{fig:DiabetesCG_GPT}, \ref{fig:DiabetesCG_Gemini}, \ref{fig:DiabetesCG_Grok-3}, and \ref{fig:DiabetesCG_LLaMa3}. These visual results offer a more comprehensive view of the causal graph structures produced by each method under different LLM backbones, thereby facilitating a detailed qualitative assessment of their structural discovery capabilities. Likewise, the corresponding qualitative comparisons on the Respiratory dataset are presented in Figures~\ref{fig:RespiratoryCG_GPT}, \ref{fig:RespiratoryCG_Gemini}, \ref{fig:RespiratoryCG_Grok-3}, and \ref{fig:RespiratoryCG_LLaMa3}.

\subsection{Full Ablation Study Results}
\label{appendix:ablation_results}

To complement the findings in Section~4.4, we further present detailed ablation results on the Diabetes and Respiratory datasets using GPT-4o as a representative LLM backbone. These results offer a more fine-grained analysis of the role of each component in the proposed framework for causal factor identification and causal structure discovery. Specifically, they allow us to examine how removing individual components affects both factor-level performance and the structural quality of the inferred causal graphs. The complete results are reported in Table~\ref{tab:FullAblation}.

Additionally, in Table~\ref{tab:AblationDesignChoices}, we study key design choices in DKCD (Figure~\ref{fig:DKCD}). SM denotes semantic matching, while SP refers to replacing the structured prompt (Prompt~\ref{prompt:latent_mining}) in knowledge-guided causal reasoning with a simpler task-only prompt that omits the reasoning chain. As shown, removing either component degrades performance, with noticeable drops in both factor identification (NP/NR/NF) and structural metrics (AR/AF/ESHD), highlighting the importance of both semantic matching and structured reasoning.

\subsection{Parameter Analysis Results}
\label{appendix:parameter}

In this section, we analyze the impact of the parameter $r$ on DKCD performance on the Diabetes dataset using GPT-4o. Here, $r$ denotes the number of top-$r$ relevant knowledge contexts selected for each factor after semantic matching. When $r$ is small, insufficient knowledge is retrieved, leading to lower performance. As shown in Table~\ref{tab:TopRStudy}, performance improves as $r$ increases and reaches a relatively good level at around $r=20$. In contrast, incorporating too much contextual knowledge can introduce excessive noise, thereby degrading DKCD performance. Therefore, we use $r=20$ in the main experiments.

\subsection{Generated Causal Clues by DKCD}
\label{appendix:causal_clue_example}
\begin{tcolorbox}[
colback=gray!10,
colframe=gray!60,
title={Example of Generated Causal Clues},
fonttitle=\bfseries,
breakable
]
\small
\textbf{Input Factors:} Lung Cancer, Tuberculosis, Lung Disease, Chest X-ray, Dyspnea, Aisa Travel.

\vspace{0.5em}
\textbf{Generated Causal Clues:}
\begin{enumerate}

    \item \textbf{Tuberculosis $\rightarrow$ Lung Disease:} Tuberculosis infection can lead to severe pulmonary damage and respiratory disease.
    
    \item \textbf{Lung Disease $\rightarrow$ Dyspnea:} Progressive lung disease is strongly associated with breathing difficulty and dyspnea.
    
    \item \textbf{Lung Disease $\rightarrow$ Chest X-ray:} Patients with suspected lung disease commonly undergo chest X-ray examinations for clinical assessment.
    
    \item \textbf{Tuberculosis $\rightarrow$ Asia Travel:} Recent travel to TB-endemic regions increases the likelihood of tuberculosis exposure.

\end{enumerate}
\end{tcolorbox}

\begin{table*}[t]
\centering
\small
\setlength{\tabcolsep}{4pt}
\caption{Full ablation study on DKCD components.}
\label{tab:FullAblation}

\resizebox{\textwidth}{!}{
\begin{tabular}{llccccccc}
\toprule
Dataset & Variant & NP$\uparrow$ & NR$\uparrow$ & NF$\uparrow$ & AP$\uparrow$ & AR$\uparrow$ & AF$\uparrow$ & ESHD$\downarrow$ \\
\midrule

\multirow{5}{*}{Diabetes dataset}
& w/o CC & \textbf{0.84 $\pm$ 0.16} & \textbf{0.71 $\pm$ 0.08} &\textbf{0.77 $\pm$ 0.04} & 0.49 $\pm$ 0.16\ & 0.31 $\pm$ 0.13 & 0.38 $\pm$ 0.15 & 26.67 $\pm$ 2.52 \\
& w/o LF  & 0.51 $\pm$ 0.17 & 0.24 $\pm$ 0.11 & 0.33 $\pm$ 0.14 & 0.00 $\pm$ 0.00 & 0.00 $\pm$ 0.00  & 0.00 $\pm$ 0.00  & 34.00 $\pm$ 1.73 \\
& w/o Both & 0.43 $\pm$ 0.14 & 0.24 $\pm$ 0.11 & 0.31 $\pm$ 0.13 & \textbf{0.67 $\pm$ 0.58} & 0.05 $\pm$ 0.05 & 0.10 $\pm$ 0.09 & 35.67 $\pm$ 3.51 \\
& DKCD & \textbf{0.84 $\pm$ 0.16} & \textbf{0.71 $\pm$ 0.08} &\textbf{0.77 $\pm$ 0.04} & 0.50 $\pm$ 0.17 & \textbf{0.33 $\pm$ 0.18} & \textbf{0.40 $\pm$ 0.18} & \textbf{25.33 $\pm$ 3.21} \\

\midrule

\multirow{5}{*}{Respiratory dataset}
& w/o CC & \textbf{0.74 $\pm$ 0.13} & \textbf{0.75 $\pm$ 0.00} & \textbf{0.74 $\pm$ 0.07} & 0.73 $\pm$ 0.14 & 0.54 $\pm$ 0.08 & 0.62 $\pm$ 0.10 & 17.00 $\pm$ 2.65 \\
& w/o LF  & 0.50 $\pm$ 0.17 & 0.42 $\pm$ 0.07 & 0.45 $\pm$ 0.11 & \textbf{1.00 $\pm$ 0.00} & 0.21 $\pm$ 0.14 & 0.32 $\pm$ 0.16 & 22.33 $\pm$ 4.51 \\
& w/o Both & 0.38 $\pm$ 0.16 & 0.30 $\pm$ 0.14 & 0.33 $\pm$ 0.15 & 0.33 $\pm$ 0.29 & 0.09 $\pm$ 0.08 & 0.14 $\pm$ 0.12 & 25.33 $\pm$ 1.53 \\
& DKCD & \textbf{0.74 $\pm$ 0.13} & \textbf{0.75 $\pm$ 0.00} & \textbf{0.74 $\pm$ 0.07} & 0.87 $\pm$ 0.23 & \textbf{0.56 $\pm$ 0.22} & \textbf{0.68 $\pm$ 0.21} & \textbf{16.67 $\pm$ 3.21} \\

\bottomrule
\end{tabular}
}
\end{table*}

\begin{table*}[t]
\centering
\small
\setlength{\tabcolsep}{4pt}
\caption{Ablation study on design choices in DKCD.}
\label{tab:AblationDesignChoices}

\resizebox{\textwidth}{!}{
\begin{tabular}{llccccccc}
\toprule
Dataset & Variant & NP$\uparrow$ & NR$\uparrow$ & NF$\uparrow$ & AP$\uparrow$ & AR$\uparrow$ & AF$\uparrow$ & ESHD$\downarrow$ \\
\midrule

\multirow{5}{*}{Diabetes dataset}
& w/o SM & 0.36 $\pm$ 0.06 & 0.19 $\pm$ 0.04 & 0.23 $\pm$ 0.05 & 0.72 $\pm$ 0.49\ & 0.08 $\pm$ 0.00 & 0.15 $\pm$ 0.01 & 34.67 $\pm$ 1.15 \\
& w/o SP  & 0.60 $\pm$ 0.06 & 0.29 $\pm$ 0.00 & 0.39 $\pm$ 0.01 & \textbf{0.78 $\pm$ 0.39} & 0.18 $\pm$ 0.09  & 0.29 $\pm$ 0.14  & 29.67 $\pm$ 2.52 \\
& DKCD & \textbf{0.84 $\pm$ 0.16} & \textbf{0.71 $\pm$ 0.08} &\textbf{0.77 $\pm$ 0.04} & 0.50 $\pm$ 0.17 & \textbf{0.33 $\pm$ 0.18} & \textbf{0.40 $\pm$ 0.18} & \textbf{25.33 $\pm$ 3.21} \\

\bottomrule
\end{tabular}
}
\end{table*}

\begin{figure*}[t]
    \centering
    \includegraphics[width=\textwidth,height=0.28\textheight,keepaspectratio]{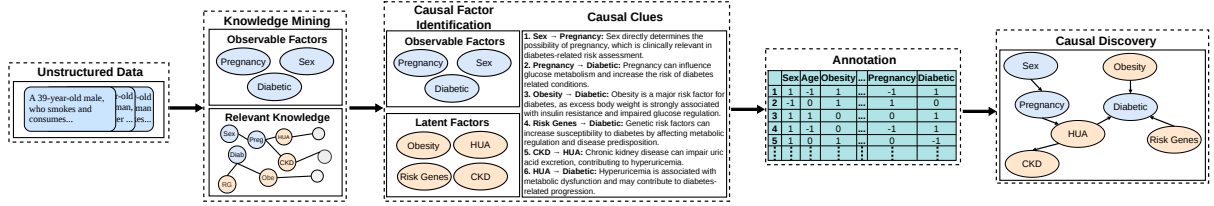}
    \caption{An example of the causal discovery process of DKCD on the Diabetes dataset.}
    \label{fig:ExampleDKCD}
\end{figure*}

\begin{table*}[t]
\centering
\small
\setlength{\tabcolsep}{4pt}
\caption{Impact of the top-$r$ relevant knowledge contexts selected during semantic matching in the knowledge mining module on the performance of DKCD.}
\label{tab:TopRStudy}

\resizebox{\textwidth}{!}{
\begin{tabular}{llccccccc}
\toprule
Dataset & $r$ & NP$\uparrow$ & NR$\uparrow$ & NF$\uparrow$ & AP$\uparrow$ & AR$\uparrow$ & AF$\uparrow$ & ESHD$\downarrow$ \\
\midrule

\multirow{4}{*}{Diabetes dataset}
& 5  & 0.33 $\pm$ 0.17 & 0.14 $\pm$ 0.07 & 0.20 $\pm$ 0.10 & 0.02 $\pm$ 0.03 & 0.03 $\pm$ 0.05 & 0.05 $\pm$ 0.08 & 36.67 $\pm$ 2.08\\
& 10 & 0.55 $\pm$ 0.15 & 0.26 $\pm$ 0.09 & 0.36 $\pm$0.11 & \textbf{0.83 $\pm$ 0.29}& 0.13 $\pm$ 0.04 & 0.21 $\pm$ 0.06 & 32.00 $\pm$ 2.65 \\
& 20 & \textbf{0.84 $\pm$ 0.16} & \textbf{0.71 $\pm$ 0.08} & \textbf{0.77 $\pm$ 0.04} & 0.50 $\pm$ 0.17 & \textbf{0.33 $\pm$ 0.18} & \textbf{0.40 $\pm$ 0.18} & \textbf{25.33 $\pm$ 3.21} \\
& 50 & 0.61 $\pm$ 0.03 & 0.41 $\pm$ 0.15 & 0.48 $\pm$ 0.11 & 0.39 $\pm$ 0.24 & 0.16 $\pm$ 0.13 & 0.22 $\pm$ 0.17 & 33.00 $\pm$ 1.00 \\

\bottomrule
\end{tabular}
}
\end{table*}

\end{document}